\documentclass[10pt,journal,compsoc]{IEEEtran}

\ifCLASSOPTIONcompsoc
  \usepackage[nocompress]{cite}
\else
  \usepackage{cite}
\fi

\newcommand{\keypoint}[1]{\noindent\textbf{#1}.\quad}

\usepackage{graphicx}
\usepackage{tabularx}
\usepackage{booktabs}
\usepackage{amsmath}
\usepackage{amsfonts}
\usepackage{xcolor}

\usepackage{hyperref}

\hypersetup{
  colorlinks = true, %
  urlcolor = blue, %
  linkcolor = red, %
  citecolor = blue %
}
\usepackage{url}

\makeatletter
\newcommand\subsubsubsection{\@startsection{paragraph}{4}{\z@}%
                                    {-3.25ex \@plus -1ex \@minus -.2ex}%
                                    {0.5ex \@plus .2ex}%
                                    {\normalfont\normalsize}}
\makeatother

\usepackage{pifont}
\newcommand{\cmark}{\ding{51}}%
\newcommand{\xmark}{\ding{55}}%

\newcommand{\cut}[1]{}

\DeclareMathOperator*{\argmin}{argmin}

\begin{document}
\title{Self-Supervised Multimodal Learning: A Survey}

\author{Yongshuo Zong, Oisin Mac Aodha, and~Timothy Hospedales,~\IEEEmembership{Senior~Member,~IEEE}%
\IEEEcompsocitemizethanks{\IEEEcompsocthanksitem Y.Z., O.M.A., and T.H. are with the School of Informatics, University of Edinburgh. T.H. is also with Samsung AI Research Centre, Cambridge.\\
E-mail: yongshuo.zong@ed.ac.uk}
}

\IEEEtitleabstractindextext{%
\begin{abstract}
Multimodal learning, which aims to understand and analyze information from multiple modalities, has achieved substantial progress in the supervised regime in recent years.  %
However, the heavy dependence on data paired with expensive human annotations impedes scaling up models. 
Meanwhile, given the availability of large-scale unannotated data in the wild, self-supervised learning has become an attractive strategy to alleviate the annotation bottleneck. 
Building on these two directions, self-supervised multimodal learning (SSML) provides ways to learn from raw multimodal data. In this survey, we provide a comprehensive review of the state-of-the-art in SSML, in which we elucidate three major challenges intrinsic to self-supervised learning with multimodal data: (1) learning representations from multimodal data without labels, (2) fusion of different modalities, and (3) learning with unaligned data. We then detail existing solutions to these challenges. Specifically, we consider: (1) objectives for learning from multimodal unlabeled data via self-supervision, (2) model architectures from the perspective of different multimodal fusion strategies, and (3) pair-free learning strategies for coarse-grained and fine-grained alignment.
We also review real-world applications of SSML algorithms in diverse fields such as control, healthcare, and remote sensing. Finally, we discuss challenges and future directions for SSML. A collection of related resources can be found at: \url{https://github.com/ys-zong/awesome-self-supervised-multimodal-learning}.
\end{abstract}

}

\maketitle

\IEEEdisplaynontitleabstractindextext

\IEEEpeerreviewmaketitle

\IEEEraisesectionheading{\section{Introduction}\label{sec:introduction}}

\IEEEPARstart{H}{umans} perceive the world through multiple senses, including sight, sound, touch, and smell. We obtain a comprehensive understanding of our surroundings by leveraging complementary information from all these modalities. 
Artificial intelligence (AI) research has long aimed to develop agents that mimic human behavior and understand the world in a similar manner. 
To this end, the field of multimodal machine learning~\cite{baltruvsaitis2018multimodal,liang2022foundations} aims to develop models that can process and integrate data from several modalities. 
In recent years, multimodal learning has made significant progress, leading to a range of applications in areas such as vision and language learning~\cite{uppal2022multimodal}, video understanding~\cite{abdu2021multimodal, chen2019deep}, biomedicine~\cite{acosta2022multimodal}, autonomous driving~\cite{huang2022multi}, etc. More fundamentally, multimodal learning is advancing the long-standing grounding problem in AI~\cite{barsalou2008grounded}, bringing us one step closer to more general-purpose AI.

However, multimodal methods often still require expensive human annotation for effective training, which impedes their scaling. %
Recently, self-supervised learning (SSL) \cite{jing2020self, ericsson2022self} has begun to alleviate  this issue by generating supervision from readily available unannotated data. The definition of self-supervision in unimodal learning is quite well established, depending solely on the training objective, and whether it exploits manual annotation or not for supervision. However, it is more nuanced in the context of multimodal learning. In multimodal learning, one modality often serves as a supervision signal for another. In terms of the goal of removing the manual annotation bottleneck for scaling up, the crucial issue defining the scope of self-supervision is then whether the cross-modal pairing is essentially freely available (e.g., as in video+audio tracks in movies), or not (e.g, as in text descriptions of images).

\begin{figure}[t]
\centering
\includegraphics[width=\linewidth]{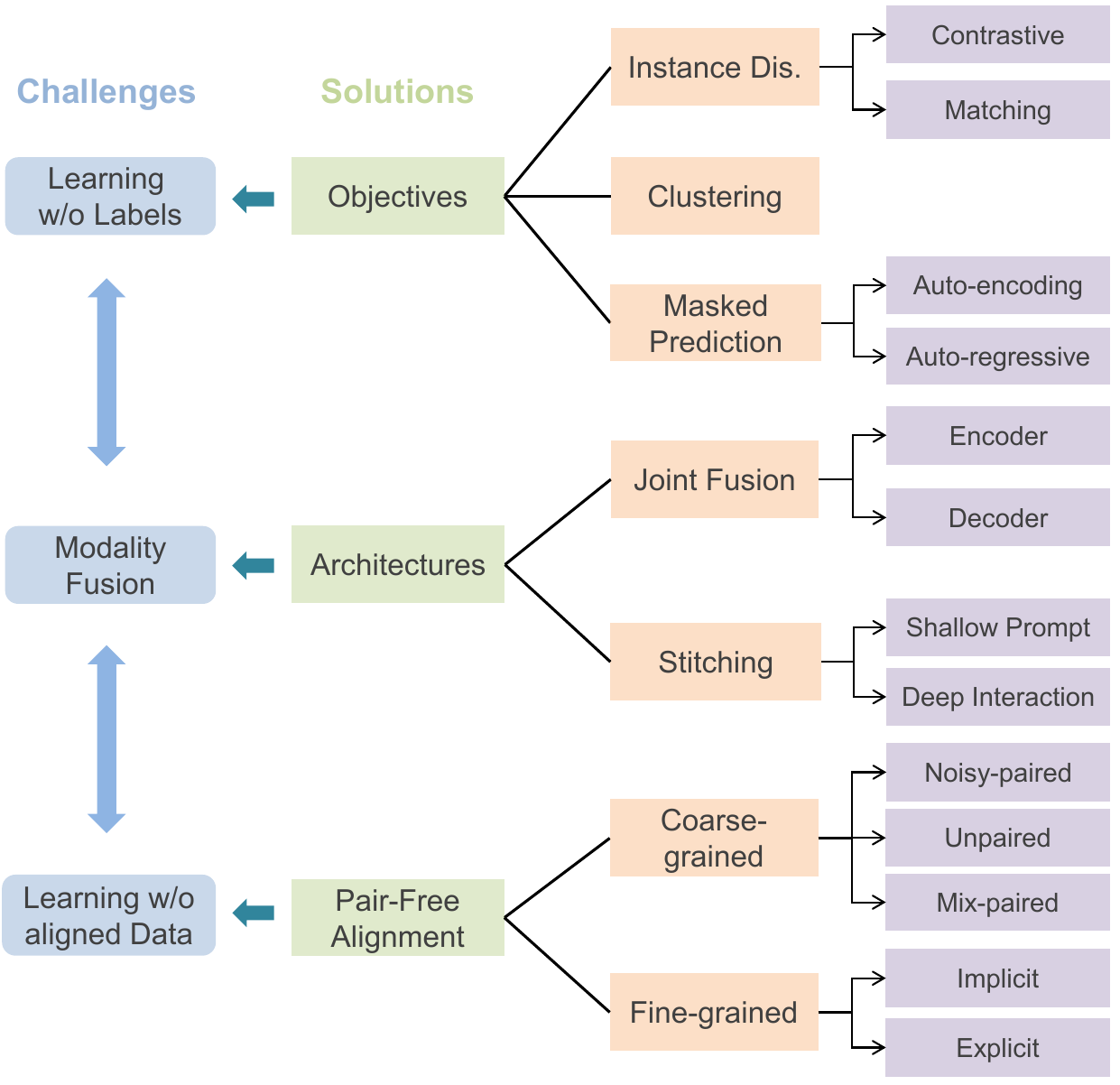}

\caption{Challenges and solutions for self-supervised multimodal learning.}
\label{fig:taxo}
\end{figure}

By exploiting freely available multimodal data and self-supervised objectives, self-supervised multimodal learning (SSML) has significantly enhanced the capability and power of multimodal models. In this survey, we review the field of SSML starting from identifying three major challenges intrinsic to the intersection of self-supervision and multi-modal data, namely: (1) \emph{how to conduct multimodal representation learning without labels?}, (2) \emph{how to fuse different modalities?}, and (3) \emph{how to learn from partially or completely unaligned multimodal data?} Starting from these inter-related questions, we discuss state-of-the-art solutions and open questions, as summarized by the taxonomy in Fig.~\ref{fig:taxo}.

To learn representation from unlabeled multimodal data, we consider different self-supervised objective functions. Based on the pretext tasks, we classify the training objectives into instance discrimination, clustering, and masked prediction categories. Hybrid methods that combine two or more of these approaches are also discussed.

Self-supervised multimodal fusion can be achieved in two ways: multimodal pretraining with a fusion architecture or amalgamation of independently pretrained unimodal models (i.e.,~``stitching''). We first scrutinize the designs of contemporary SSML architectures for multimodal pre-training. Specifically, we consider the design space for encoder and fusion modules, contrasting modality-specific encoders (without fusion or with late fusion) and unified encoders with early fusion. We also examine specific decoder designs, and discuss the impact of these design choices. In contrast, model stitching refers to linking two or more separately pretrained unimodal models via self-supervision, thereby stitching together a multimodal model.

The final unique challenge is learning with unaligned data, which is key for self-supervision as the alignment itself is often a crucial missing annotation in multi-modal data. Multi-modal data may be (un)aligned at two different levels: coarse-grained (e.g., image-caption) and fine-grained (e.g., bounding box-word). We outline strategies for pair-free learning that aim to accomplish alignment at these levels. For coarse-grained data, we discuss approaches for learning from various pairing scenarios, including noisy-paired, mixed-paired, and completely unpaired data. On the other hand, for fine-grained data, our focus is on methodologies to derive implicit or explicit fine-grained alignment.

Finally, we discuss the applications of these algorithms in various real-world areas, including healthcare, remote sensing, machine translation, etc., and provide in-depth discussions of the technical challenges and societal impact of SSML, highlighting potential future research directions. We summarize the latest advances in methods, datasets, and implementations to provide a starting point for researchers and practitioners in this field.

Existing survey papers either focus solely on \emph{supervised} multimodal learning \cite{baltruvsaitis2018multimodal, liang2022foundations, ramachandram2017deep, xu2022multimodal} or \emph{unimodal} self-supervised learning \cite{jing2020self, ericsson2022self, liu2021self}, or specific subareas of SSML, e.g., vision-language pretraining \cite{gan2022vision}. 
The most relevant review is \cite{deldari2022beyond}, but it focuses mostly on temporal data and omits the key considerations of fusion and alignment in SSML. In contrast, we provide a comprehensive and up-to-date review of SSML algorithms with a new taxonomy that covers unique challenges and solutions in SSML. 

\section{Background}\label{sec:problem}
We first introduce the notation used in this survey, and then formally introduce the learning paradigms. Finally, we define the scope of SSML algorithms we will cover.

\subsection{Notation}
\label{sec:notations}
\keypoint{Unimodal and Multimodal Data}
A modality is a category of data defined by how it is received, represented, and understood. Unimodal data refers to data from only one modality, e.g., images. A unimodal dataset $\mathcal{D}_u = \{(x^1, y^1), (x^2, y^2), \ldots, (x^n, y^n)\}$ consists of inputs $x$ and optionally ground-truth labels $y$, where $n$ is the total number of data points. In contrast, multimodal data is from different ($\geq 2$) modalities, e.g., images and text, which may contain shared or complementary information. A multimodal dataset can be represented as $\mathcal{D}_m = \{(x^1_1, \ldots, x^1_k, y^1), \ldots, (x^n_1, \ldots, x^n_k, y^n)\}$, where $k$ is the number of input modalities. This is \emph{paired} multimodal data, because samples from each modality are paired into tuples $(x^i_1, \ldots, x^i_k)$ where observations refer to the same real-world event, or otherwise correspond.

In some cases where paired multimodal data is difficult to obtain, unpaired data can also be used by designing specific algorithms. \textit{Unpaired} multimodal data refers to the datasets where each modality is recorded independently, without producing corresponding pairs of observations from each input modality. 
Formally, for a dataset $\mathcal{D}_m$ with $K$ modalities and $N$ samples, we can divide the samples into $K$ separate datasets $\{\mathcal{D}_k = \{x^i_k, y^i_k\}_{i=1}^{n_k}\}_{k=1}^K$, where $\sum_{k=1}^{K} n_k=N$. Furthermore, \textit{mixed} multimodal data can be considered as a combination of paired and unpaired data. 

\keypoint{Unimodal and Multimodal Models} 
Without loss of generality, consider we want to solve a predictive task. For unimodal learning, the goal is to learn a predictive model $h$, where $h$ usually consists of a representation encoder $e_\theta$ and a task-specific predictive head $g_\phi$, i.e., $\hat{y}=h(x)=g_\phi(e_\theta(x))$. For multimodal learning, we also want to learn a predictive model $h$. In contrast, $h$ is usually composed of modality-specific encoder $e_{\theta_k}$ for each input $x_k$, a potential fusion module $f_\psi$ to integrate the encoded information of different modalities, and a predictive head $g_\phi$. For simplicity, we denote $e_{\theta}(x_1, \ldots, x_k) = (e_{\theta_1}(x_1), \ldots, e_{\theta_k}(x_k))$ in the following text, i.e., $\hat{y}=h(x)=g_\phi(f_\psi(e_{\theta}(x_1, \ldots, x_k))$. $\phi$, $\theta$, and $\psi$ are the parameters of the corresponding models. The specific architecture designs are discussed in Section \ref{sec:architecture}. The parameters are optimized by a loss function $\mathcal{L}$, and the learning procedures are detailed in Section \ref{sec:define_ssml}.

\subsection{Learning Paradigms}
\label{sec:define_ssml}
\begin{figure}[t]
\centering
\includegraphics[width=\linewidth]{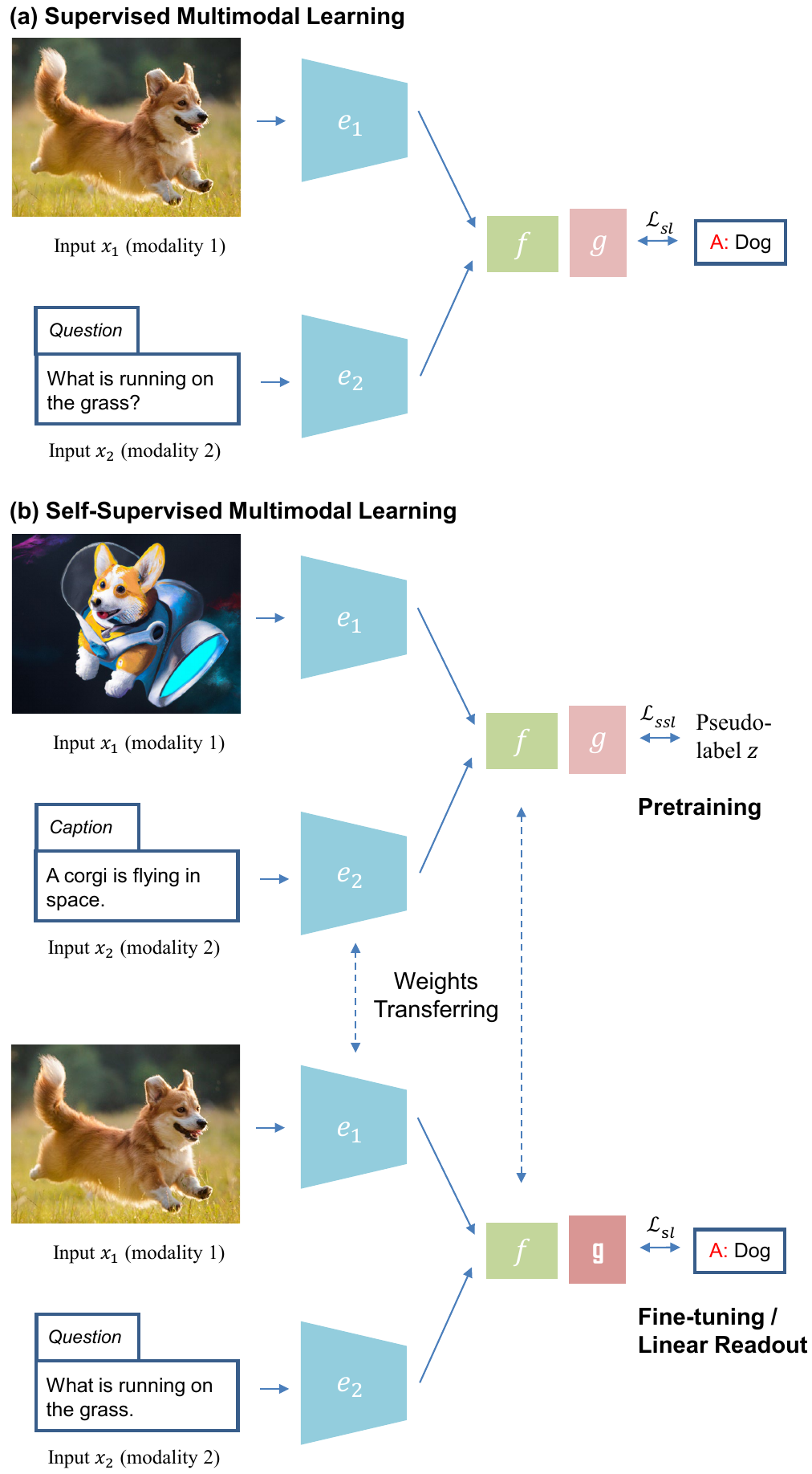}

\caption{Learning paradigms for (a) supervised multimodal learning, and (b) self-supervised multimodal learning; illustrating self-supervised pretraining without manual annotations (top) and supervised fine-tuning or linear readout for downstream tasks (bottom).}
\label{fig:paradigms}
\end{figure}

Here we summarize the four related machine learning paradigms: supervised unimodal learning, self-supervised unimodal learning, supervised multimodal learning, and self-supervised multimodal learning. We illustrate the multimodal learning paradigms in Fig.~\ref{fig:paradigms}.

\keypoint{Supervised Unimodal Learning}
Supervised unimodal learning aims to learn the predictive model $h$ by using the supervision of the human-annotated ground-truth labels $y$. The parameters of the encoder $e_\theta$ and the predictive head $g_\phi$ are optimized through a supervised loss function $\mathcal{L}_{sl}$:
\begin{equation}
    \phi^*, \theta^* = \argmin_{\phi, \theta} \sum_{(x^i, y^i)\in \mathcal{D}_u} \mathcal{L}_{sl}(g_\phi(e_\theta(x^i)), y^i).
\end{equation}
Supervised unimodal learning has already revolutionized many areas and achieved human-level recognition ability, such as in computer vision \cite{he2016deep}. However, successful training of models usually requires a large number of ground-truth labels, which are not always available.

\keypoint{Self-supervised Unimodal Learning}
Contrary to supervised learning, self-supervised unimodal learning does not require the ground-truth labels $y$ during pretraining, which alleviates the requirement for expensive human annotations. Instead, by applying a pretext process $P$ to the input $x$, a pseudo-label $z = P(x)$ is generated to supervise training, e.g.,  \cite{noroozi2016unsupervised}. Similarly, the parameters of the encoder $e_\theta$ and the pretext head $g_\phi$ are optimized through a self-supervised loss function $\mathcal{L}_{ssl}$:
\begin{equation}
    \phi^*, \theta^* = \argmin_{\phi, \theta} \sum_{x^i\in \mathcal{D}_u} \mathcal{L}_{ssl}(g_\phi(e_\theta(x^i)), P(x^i)).
\end{equation}
When transferring to the downstream tasks, the pretext head $g_\phi$ is discarded and the encoder $e_\theta$ is kept as a partial solution to solve the target problem with a new task-specific head $\mathfrak{g}_\gamma$. They can be effectively trained on a small labeled dataset $D_t$ by either \textit{fine-tuning} to update the whole model or \textit{linear readout} where the weights of the encoder $e_\theta$ are frozen and only the task-specific head $\mathfrak{g}_\gamma$ is trained with a supervised loss. Formally, for fine-tuning:
\begin{equation}
    \gamma^*, \theta^* = \argmin_{\gamma, \theta} \sum_{(x^i, y^i)\in \mathcal{D}_t} \mathcal{L}_{sl}(\mathfrak{g}_\gamma(e_\theta(x^i)), y^i).
\end{equation}
And for linear readout:
\begin{equation}
    \gamma^* = \argmin_{\gamma} \sum_{(x^i, y^i)\in \mathcal{D}_t} \mathcal{L}_{sl}(\mathfrak{g}_\gamma(e_\theta(x^i)), y^i).
\end{equation}

\keypoint{Supervised Multimodal Learning}
Supervised multimodal learning follows a similar learning paradigm as supervised unimodal learning. The ground-truth labels $y$ are used to supervise the optimization of the parameters of the modality-specific encoder $e_{\theta_i}$, fusion module $f_\psi$, and the predictive head $g_\phi$:
\begin{equation}
    \phi^*, \theta^*, \psi^* =  \argmin_{\phi, \theta, \psi} \!\!
     \sum_{(x^i_k, y^i_k)\in \mathcal{D}_m} \!\!\! \mathcal{L}_{sl}(g_\phi(f_\psi(e_{\theta}(x^i_1, \ldots, x^i_k)), y^i)).
\end{equation}
An example of a supervised multimodal learning task is visual question answering, as illustrated in Fig.~\ref{fig:paradigms}~(a). 

\keypoint{Self-supervised Multimodal Learning}
Similar to self-supervised unimodal learning, the ground-truth label $y$ is not available during pretraining. The pseudo-label $z$
can be generated by any of the input modalities $z=P(x_j)$
or jointly from some or all input modalities (e.g., $z=P(x_i, x_j)$). %
For generality, we denote the pseudo-label transformation for self-supervised multimodal learning as $z=P(x_1, \ldots, x_k)$. Then, the parameters of the encoder $e_{\theta}$, fusion module $f_\psi$, and the predictive head $g_\phi$ can be trained by minimizing a self-supervised loss $\mathcal{L}_{ssl}$:
\begin{equation}
\label{eq:ssml}
    \phi^*, \theta^*, \psi^* = \argmin_{\phi, \theta, \psi} \!\!
    \sum_{(x^i_k)\in \mathcal{D}_m} \!\!\! \mathcal{L}_{ssl}(g_\phi(f_\psi(e_{\theta}(x^i_1, \ldots, (x^i_k)), z^i)).
\end{equation}
Following pretraining, the downstream transfer process is analogous to that of the downstream unimodal case. With the task-specific head $\mathfrak{g}_\gamma$, we can fine-tune as:
\begin{equation}
\label{eq:mmft}
    \gamma^*, \theta^*, \psi^* = \argmin_{\gamma, \theta, \psi} \!\!
    \sum_{(x^i_k, y^i_k)\in \mathcal{D}_t} \!\!\! \mathcal{L}_{sl}(\mathfrak{g}_\gamma(f_\psi(e_{\theta}(x^i_1, \ldots, (x^i_k)), y^i)).
\end{equation}
Alternatively, for linear readout, we can freeze the encoder $e_{\theta}$ and the fusion model $f_\psi$:
\begin{equation}
\label{eq:mmlr}
    \gamma^* = \argmin_{\gamma} \!\!
    \sum_{(x^i_k, y^i_k)\in \mathcal{D}_t} \!\!\! \mathcal{L}_{sl}(\mathfrak{g}_\gamma(f_\psi(e_{\theta}(x^i_1, \ldots, (x^i_k)), y^i)).
\end{equation}
An example illustrating the self-supervised vision and language pretraining prior to downstream supervised learning for visual question answering is shown in Fig.~\ref{fig:paradigms}~(b).

\subsection{Scope of the Survey}
\label{sec:scope}
\subsubsection{Self-supervision in Multimodal Learning}
We first delineate the scope of SSML as considered in this survey as this term has been used inconsistently in the literature before. Self-supervision is more straightforward to define in a unimodal context by appealing to the label-free nature of different pretext tasks, e.g., as famously realized by instance discrimination~\cite{chen2020simple} or masked prediction objectives~\cite{He2021MaskedAA}. In contrast, the situation in multimodal learning is more complex as the role of modalities and labels becomes blurred. For example,  text is typically considered a label in supervised image captioning~\cite{Lin2014MicrosoftCC}, but as an input modality in self-supervised multimodal vision and language representation learning~\cite{radford2021learning}. 

In the multimodal context, the term self-supervision has been used to refer to at least four situations: (1) Label-free learning from multimodal data that is automatically paired -- such as movies with video and audio tracks~\cite{Arandjelovi2017LookLA}, or images and depth data from RGBD cameras~\cite{Tian2019ContrastiveMC}. 
(2) Learning from multimodal data where one modality has been manually annotated, or two modalities have been manually paired, but this annotation has already been created for a different purpose, and can thus be considered free for the purpose of SSML pretraining. For example, matching image-caption pairs crawled from the web, as used by the seminal CLIP~\cite{radford2021learning}, is actually an example of supervised metric-learning~\cite{Chopra2005LearningAS,lu2017deepMetric} where the pairing is the supervision. However, because both the modalities and the pairing are all freely available at scale, it is often described as self-supervised. Such uncurated incidentally created data is often lower-quality and noisier than purpose-created and curated datasets such as COCO~\cite{Lin2014MicrosoftCC} and Visual Genome~\cite{Krishna2016VisualGC}. 
(3) Learning from high-quality purpose annotated multimodal data (e.g., manually captioned images in the COCO~\cite{Lin2014MicrosoftCC}), but with self-supervised style objectives, e.g., Pixel-BERT~\cite{Huang2020PixelBERTAI}. 
(4) Finally, there are `self-supervised' methods that use a mix of free and manually annotated multimodal data~\cite{Kim2021ViLTVT, Wang2022ImageAA}. 

For the purpose of this survey, we follow the spirit of self-supervision as aiming to scale up by breaking the manual annotation bottleneck. Thus we include methods that fall into the first two and fourth categories above in terms of being able to train on freely available data. We exclude methods only shown to work on manually curated datasets just because they apply typically ``self-supervised'' objectives (e.g., masked prediction) on curated datasets.

\subsubsection{Multimodal v.s. Multiview}
Multimodal learning and multiview learning are related, but distinct, concepts that are used interchangeably in some literature \cite{xu2022multi, xu2022self}. Both aim to extract complementary information from more than one data source to improve performance at a task. However they differ in that multimodal learning focuses on learning from data that originates from multiple heterogeneous modalities of data, such as text, image, audio, or gene sequences (e.g., \cite{Srivastava2012MultimodalLW, ramesh2021zero, Li2019VisualBERTAS}). On the other hand, multiview learning deals with multiple views of data obtained from the same modality. For example, photos of an object from different viewpoints or audio recordings from spatially offset microphones (e.g., \cite{Wang2015OnDM, hassani2020contrastive, Wang2017UnsupervisedMF}). Multiview learning also spans situations where different features are extracted from a single modality of observation -- such as amplitude and phase after the Fourier transform of an input. In this survey, we focus on multimodal learning only, where inputs are distinct heterogeneous modalities.

\subsubsection{Generative vs Self-Supervised Models}
Generative models, such as generative adversarial networks (GANs) \cite{GAN2014} and diffusion models \cite{ho2020DDPM}, are unsupervised learners. We primarily focus on self-supervised algorithms for representation learning. Thus, we will exclude discussion of these models in this survey and instead refer readers to other surveys \cite{zhan2021multimodal,Frolov2021AdversarialTS,Yang2022DiffusionMA} that focus on these topics.

\section{Multimodal Learning without Labels}\label{sec:Objective}
The challenge of learning multimodal representations from unlabeled data calls for bespoke solutions. The lack of labels requires learning strategies that can extract meaningful representations from different modalities without having access to explicit supervised label information. By leveraging self-supervised multimodal objectives, we can harness the inherent structure and co-occurrence patterns within the data to drive the learning process. Unlike unimodal SSL, multimodal objectives must account for different modalities and their inherent characteristics. This aspect distinguishes it as a distinct challenge, necessitating specialized solutions. Specifically, we detail objective functions designed for training self-supervised multimodal algorithms across three primary categories: instance discrimination, clustering, and masked prediction, focusing on how classic unimodal objectives can be extended to handle multimodal inputs. In addition, we also discuss hybrid objectives at the end.

\subsection{Instance Discrimination}
In unimodal learning, instance discrimination (ID) treats each instance in the raw data as a separate class, and the model is trained to differentiate between different instances. In the context of multimodal learning, instance discrimination often aims to determine whether samples from two input modalities are from the same instance, i.e., paired. By doing so, it attempts to align the representation space of the paired modalities while pushing the representation space of different instance pairs further apart. There are two types of instance discrimination objectives: contrastive and matching prediction, depending on how the input is sampled.

\begin{figure}[t]
\centering
\includegraphics[width=\linewidth]{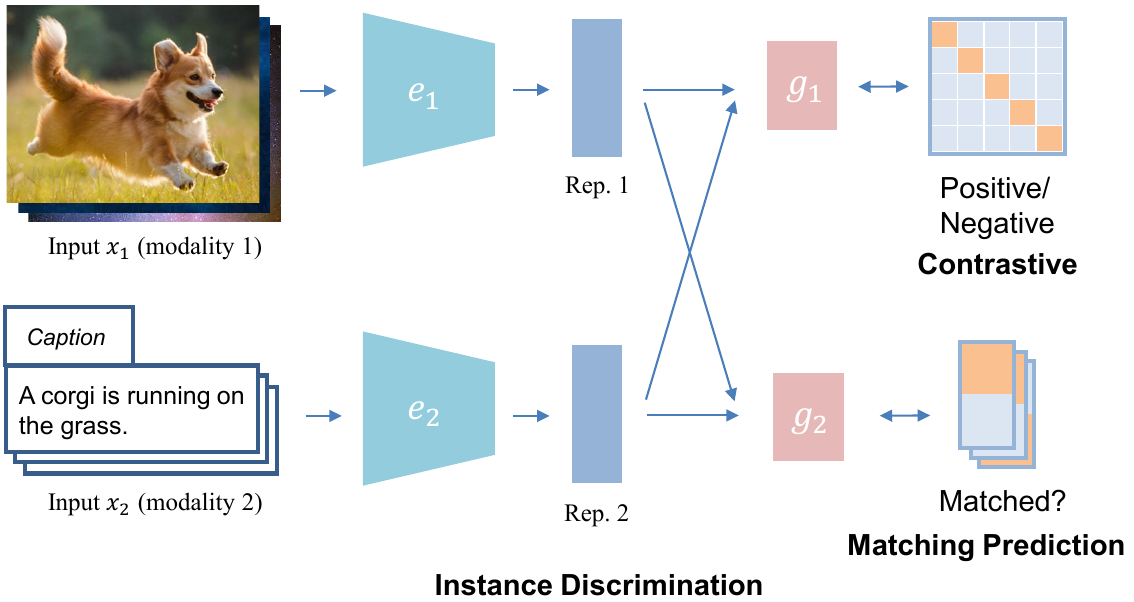}

\caption{An illustrative schematic of instance discrimination objectives.}
\label{fig:contrastive}
\end{figure}

\subsubsection{Contrastive}
Contrastive methods typically use corresponding samples from different modalities as positive example pairs and non-corresponding samples as negative pairs. These pairs are then used to train a model to accurately distinguish positive and negative pairs using contrastive training objectives.

Given an anchor data point $x^{a}$ drawn from modality $k$, the other modality from the same instance is selected as positive sample $x^{+}$, and non-corresponding points are regarded as negative samples $x^{-}$. After encoding, the extracted representations for anchor, positive, and negative samples can be defined as $r^a=e(x^a)$, $r^+=e(x^+)$, and  $r^-=e(x^{-})$ in feature space $F$. Then, a general form of contrastive objective can be written as:
\begin{equation}
\begin{split}
    \mathcal{L}_{Con} = 
    \sum_{(r^a,r^+,r^-)\in F}
    \left[ \log \frac{sim(r^a, r^+)}{sim(r^a, r^+)+ \sum_{j=1}^m sim(r^a, r_j^-)} \right],
\end{split}
\end{equation}
where $sim(,)$ is a similarity function between two inputs, and $m$ is the number of negative samples.

Contrastive Multiview Coding (CMC) \cite{Tian2019ContrastiveMC} is one of the earliest works to explore the application of contrastive learning in the multimodal setting. This framework maximizes the mutual information between representations of different views (modalities) of the same scene while pushing apart the unmatched samples. %
The idea of maximizing mutual information among different modalities and performing cross-modal instance discrimination has been further developed and extended in various ways. AVTS \cite{Korbar2018CooperativeLO} considers temporally synchronized audio-video pairs as positives and utilizes curriculum Learning to gradually learn hard negatives. To achieve spatial alignment, AVSA \cite{morgado2020learning} samples video and audio clips from different spatial viewing directions and maximizes the mutual information of audio-visual pairs in the same direction. MultiModal Versatile (MMV) networks \cite{alayrac2020self} maximize the mutual information among vision, audio, and text pairs that temporally co-occur. %
Video-Audio-Text transformer (VATT) \cite{akbari2021vatt} uses a similar contrastive objective, but studies a modality-agnostic, single-backbone transformer by sharing weights among the three modalities.

Contrastive learning has also shown great potential for scaling up to larger models and datasets. 
CLIP \cite{radford2021learning} achieves strong zero-shot performance when pretraining on a large dataset of 400 million image-language pairs, by simply predicting the pairing. This  paradigm has been successfully adopted in various other domains, such as AudioCLIP \cite{Guzhov2021AudioclipEC}, VideoCLIP \cite{Xu2021VideoCLIPCP}, CLIP4CLIP \cite{Luo2021CLIP4ClipAE}, and pointCLIP \cite{Zhang2021PointCLIPPC}. As collecting paired data may require additional curation steps, ALIGN \cite{jia2021scaling} studies whether pretraining on noisy pairs still achieves strong performance and confirms that it does.
As the model and dataset size increase, training becomes more computationally expensive and methods are developed to improve efficiency. %
CLIPPO~\cite{Tschannen2022ImageandLanguageUF} unifies CLIP modalities using a pixel-based approach for image, text, and multimodal tasks, while FLIP~\cite{li2022flip} enhances training efficiency by masking out large sets of image patches, both yielding performance on par with or superior to CLIP.

Besides inter-modal contrastive learning for alignment, conventional intra-modality learning can provide an additional cue. %
SLIP \cite{mu2022slip} and CrossCLR \cite{zolfaghari2021crossclr} add intra-modal contrastive losses alongside cross-modal learning leading to improved performance in image-text and video-text problems.
CrossPoint \cite{Afham2022CrossPointSC} learns cross-modal correspondence from point clouds and rendered images and intra-modal correspondence from different views of a 3D point cloud.
However, the addition of within-modality instance discrimination is not always beneficial. For example, AVID \cite{Morgado2020AudioVisualID} shows that naively adding intra-modality instance discrimination may harm the overall performance as it is an easier pretext task compared to cross-modal discrimination and can be partially solved by matching low-level data statistics.

\subsubsection{Matching Prediction}

Matching prediction, also known as alignment prediction, aims to predict whether a pair of samples from two input modalities are matched (positive pair) or not (negative pair). For example, if a piece of text corresponds to an image's caption. 
A key difference between contrastive learning and matching prediction is that in a mini-batch, the former calculates the similarity between a positive pair and all of the other negative pairs, while the latter labels individual tuples as positive or negative. Denoting $p$ to be the two-class probability of a matched pair, the matching prediction loss minimizes a binary cross-entropy loss (BCE):
\begin{equation}
\begin{split}
    & \mathcal{L}_{Match} =  \frac{1}{n} \sum_{(x^i_k)\in \mathcal{D}_m} \mathcal{L}_{BCE}(
     p^i, z^i), \\
     & \text{and } p^i=g_\phi(f_\psi(e_{\theta}(x^i_1, \ldots, x^i_k))),
\end{split}
\end{equation}
where the pseudo-label $z^i$ is a one-hot vector representing whether the inputs are matched.

Matching prediction is widely used for modeling audio-visual correspondence (AVC). AVC was introduced by L$^3$-Net \cite{Arandjelovi2017LookLA}, which uses a fused representation from audio and video to make a binary prediction of whether the audio-image pair is from the same video clip.
This strategy is adopted by AVE-Net \cite{Arandjelovi2017ObjectsTS} by only using Euclidean distance alignment without fusion, leading to localization of the object that sounds within an image. Owens and Efros \cite{Owens2018AudioVisualSA} also utilize this pretext task, but instead take temporal video frames and audio as input. They construct negative pairs from the same video to increase pretext task difficulty, improving learned representations and localization accuracy.

In order to achieve better audio-visual localization and separation, pixel-wise matching prediction has been proposed as a pretext task. One such example is the mix-and-separate method \cite{Zhao2018TheSO}, which combines audio signals from different videos to create an input mixture. The network is trained to separate the audio sources by predicting binary spectrogram masks based on corresponding video frames. Similar approaches are also proposed in \cite{zhou2022audio, hu2021class}. Building on this idea, Sound of Motions \cite{Zhao2019TheSO} incorporates motion trajectory modeling, while Music Gesture \cite{Gan2020MusicGF} uses human body and hand movements to guide the separation.

Image-text matching (ITM) is an effective objective for vision-language pretraining first proposed by UNITER \cite{Chen2019UNITERUI} before CLIP. It fed the global cross-modal representation to a binary classifier to predict whether the input pair is matched. ITM has been adopted by various algorithms, including ViLBERT \cite{Lu2019ViLBERTPT}, BLIP \cite{Li2022BLIPBL}, FLAVA \cite{Singh2021FLAVAAF}, etc. ITM can also be complementary to contrastive learning. For example, ALBEF \cite{Li2021AlignBF} samples hard negatives from the contrastive batch in order to train the ITM objective with more informative negatives. 

\keypoint{Discussion} 
Instance discrimination has emerged as an effective and versatile framework for learning representations from multiple modalities. %
A key aspect of many multimodal instance discrimination methods is the strategy for sampling positive and negative samples across modalities, which can significantly impact the learned representation. 
For example, two non-corresponding samples will be treated as a negative pair regardless of their semantic similarity. This process can produce both false positive and false negative pairs thus introducing label noise \cite{chuang2020debiased, saunshi2019theoretical}. It also means that most negatives will be very easy to distinguish, leading to hard negative mining being a topic of ongoing study \cite{robinson2021contrastive, kalantidis2020hard}. 
While contrastive learning is effective, it often requires a large batch size to obtain enough negative samples to avoid mode collapse. This is resource intensive, especially in terms of memory, and has led to active research in more efficient contrastive learning \cite{bordes2023towards}. 
When it comes to modeling correspondence or interaction between modalities, contrastive models typically take cross-modal dot products after obtaining embeddings from modality-specific encoders. This has the advantage of simplicity, but lacks the ability to model rich interactions between the modalities. 
On the other hand, matching prediction 
can be carried out on a joint representation of both modalities.
The latter approach enables richer cross-modal interactions. Thus, these two objectives are sometimes combined to achieve a complementary effect.

\subsection{Clustering}
Clustering methods assume that applying end-to-end trained clustering will lead to the grouping of the data by semantically salient characteristics. In practice, these methods iteratively predict the cluster assignments of the encoded representation, and use these predictions, also known as pseudo labels, as supervision signals to update the feature representation. Multimodal clustering provides the opportunity to learn multimodal representations and also improve conventional clustering by using each modality's pseudo-labels to supervise the other. 

Formally, let the predictive head $g$ be a standard clustering method such as K-means (i.e., with no learnable parameters). 
It clusters the encoded representations into $M$ distinct clusters based on geometric similarity. We denote the cluster prediction to be $C^i$ for input $x^i$. Clustering-based methods minimize a cross-entropy loss between the predicted cluster assignments $C^i$ and the pseudo-labels $z^i$. Proposed by DeepCluster \cite{caron2018deep}, a widely used approach to generate pseudo-labels is to jointly learn a $d \times M$ centroid matrix $T$ and the cluster assignments $z^i$ of each data sample $x^i$ by optimizing the following objective: 
\begin{equation}
\begin{split}
    \min_{T \in \mathbb{R}^{d \times M}} \frac{1}{n} \sum_{i=1}^{n} & \min_{z^i \in \{0, 1\}^k} \lVert f_\psi(e_{\theta}(x^i_1, \ldots, x^i_k)) - Tz^i \rVert_2^2 \\
    & \text{s.t. } z_i^{\top}1_k=1.
\end{split}
\end{equation}
This results in a set of optimal assignments $(z^{i*})_{n \leq N}$ and a centroid matrix $T^*$. We use the assignment as the pseudo-labels, while the centroid matrix is discarded. After that, the model can be optimized using:
\begin{equation}
\label{eq:cluster}
\begin{split}
    & \mathcal{L}_{Clustering} =  \frac{1}{n} \sum_{(x^i_k) \in \mathcal{D}_m} \mathcal{L}_{CE}(
     C^i, z^i), \\
     & \text{and } C^i=g(f_\psi(e_{\theta}(x^i_1, \ldots, x^i_k))).
\end{split}
\end{equation}
Then, the clustering process is repeated on the updated representations, and thus the model can be updated iteratively.

Cross-Modal Deep Clustering (XDC) \cite{Alwassel2019SelfSupervisedLB} is a representative clustering-based method for video and audio representation learning. It uses pseudo-labels of the cluster assignments of one modality to supervise the training of the other modality.
The authors also explore Multi-Head Deep Clustering (MDC) and Concatenation Deep Clustering (CDC), which use cluster assignments from both modalities and joint representations of both modalities as supervision, respectively. All three methods yield representations that achieve good performance on various downstream tasks.

SeLaVi \cite{Asano2020LabellingUV} builds on a unimodal SeLa \cite{Asano2019SelflabellingVS}, that learns clustering by solving an optimal transport problem, for video labeling. SeLaVi extends it to multimodal data by considering the extracted audio and visual information as different views and then learns view invariance. Non-degenerate clustering is ensured via optimal transport. DMC \cite{Hu2018DeepMC} encodes images and audio spectrograms into separate representations, which are then co-clustered. %
The model then uses the similarity across modalities as supervision for training.

Alternative pretext tasks for clustering have also been designed. AV-HuBERT \cite{shi2021learning} uses the encoded masked audio and image-sequence representations to predict a pre-determined sequence of discrete cluster assignments, termed masked cluster prediction. It is more resilient to bad cluster assignments compared to unmasked cluster prediction. u-HuBERT \cite{hsu2022u} generalizes AV-HuBERT to be compatible with both multimodal and unimodal speech by mapping various inputs to a shared modality-agnostic embedding space for masked cluster prediction.

\keypoint{Discussion} 
Clustering objectives enable the model to  capture the underlying structure of the data by using cluster assignments as supervision signals, providing strong performance on downstream tasks such as cross-modal retrieval. The cluster assignments can be generated in different ways, such as from the jointly fused representation or from modality-specific representations. Different from unimodal clustering that enforces different views of the same instance to have the same cluster assignments, we may want to allow the different modalities to have different cluster assignments to increase diversity as the paired modalities may not be perfectly matched. But it is hard to know apriori how much flexibility is optimal to have for a given dataset. Similarly, for noisy paired datasets, clustering can alleviate the issue of false positives and hard negatives that contrastive learning suffers from. However, disadvantages include sensitivity to parameter initialization, the choice of clustering algorithm, and the number of clusters chosen -- which must be carefully chosen to balance overfitting and underfitting.%

\subsection{Masked Prediction}

The masked prediction task can be either performed in an auto-encoding (similar to BERT \cite{devlin2018bert}) or an auto-regressive approach (similar to GPT \cite{radford2018improving}). 

\subsubsection{Auto-encoding Masked Prediction}
Auto-encoding masked predictors pretrain models by providing them with input data where some elements have been randomly masked, and training the model to predict the missing information. This aims to force the model to understand the relationships between different pieces of data, thus learning rich semantic features. This approach was first introduced for natural language processing with the masked language modeling (MLM) technique proposed by BERT \cite{devlin2018bert}, and is now widely used for multimodal tasks.

\begin{figure}[t]
\centering
\includegraphics[width=\linewidth]{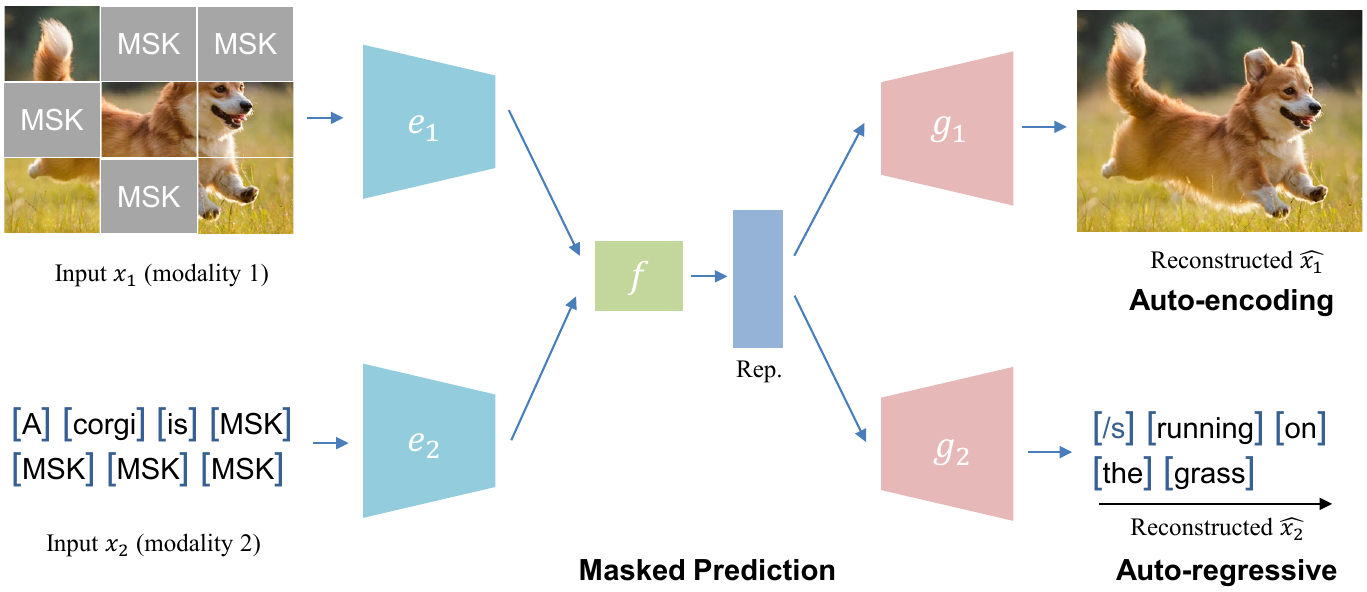}

\caption{An illustrative schematic of masked prediction frameworks.}
\label{fig:MP}
\end{figure}

For multimodal learning, the masked prediction task is often used in a cross-modal context, where the model %
should predict missing information conditioned on other modalities, as illustrated in Fig.~\ref{fig:MP}. 
For example, the model may be given an image as context and be trained to predict missing text, or vice versa. 
This requires the model to understand the relationship and interaction between different modalities. Without loss of generality, consider a pretext task that aims to generate a reconstructed input modality $x_k$ conditioning on the other modality $x_j, \text{(} j \neq k \text{)}$, where we apply a  masking function to the input, i.e., $\tilde{x} = \text{MASK}(x)$:
\begin{equation}
\label{eq:con}
    \mathcal{L}_{\text{AE}} = \frac{1}{n} \sum_{(x^i_k) \in \mathcal{D}_m} \mathcal{L}_{recon}(g_\phi(f_\psi(e_{\theta}(\tilde{x}^i_k, \tilde{x}^i_j))), x^i_k),
\end{equation}
where $\mathcal{L}_{recon}$ is usually a cross-entropy or mean squared error (MSE) loss to measure the difference between the original input and the reconstructed output. 

In some cases, intra-modal masked prediction is complementary to cross-modal masked prediction, i.e., the model must also predict intra-modal masked content based solely on the information contained within the same modality. This can help to learn representations that are robust to the absence of other modalities \cite{Bao2022VLBEiTGV, Sun2019VideoBERTAJ}. 

SelfDoc \cite{Li2021SelfDocSD}, proposed for document image understanding, introduces a masking function to randomly mask language or vision features for prediction, helping it to infer contextual clues and multimodal information.
VideoBERT \cite{Sun2019VideoBERTAJ}, designed for video and text, transforms both raw inputs into discrete token sequences, allowing them to be used by the same language model. The model can be pretrained on video-only or text-only corpora using intra-modal mask-completion objectives. VL-BEIT \cite{Bao2022VLBEiTGV} uses a shared transformer to perform masked prediction on both unimodal and multimodal data.
It makes use of a simple and effective design that can be learned from scratch with one unified pretraining task, one shared backbone, and one-stage training. BEiT-3 \cite{Wang2022ImageAA} further demonstrates that using masked prediction objectives alone achieves state-of-the-art transfer performance on various downstream tasks. It treats images as a language and performs masked ``language'' modeling on images, text, and image-text pairs in a unified manner.
Unified-IO \cite{Lu2022UnifiedIOAU} conducts masked language modeling and masked image modeling, which can be trained separately, or together when multiple modalities are present.

Besides performing exact reconstructions, methods have been proposed to instead match distributions over items in order to better model the high-level features. For example, ViLBERT \cite{Lu2019ViLBERTPT} predicts a distribution over semantic classes of the masked text input for the corresponding image region, where the target distribution is obtained by a pretrained object detector. The KL divergence is minimized between the two distributions. Similarly, ActBERT \cite{Zhu2020ActBERTLG} adopts this distribution matching strategy to model video and text.

\subsubsection{Auto-regressive Masked Prediction}
The auto-regressive pretraining method, popularized by PixelCNN \cite{van2016conditional} and GPT \cite{radford2018improving}, makes predictions of the next (masked) token one step at a time, from left to right. Formally, considering the specific modality input $x_k$ is tokenized by the encoder $e_{\theta k}$ to a set of tokens $\mathcal{U}_k =\{u^1_k, \ldots, u^n_k \}$ (e.g., a set of words in text, or set of patches in images), and similarly for input modality $x_j$, we have tokens $\mathcal{U}_j =\{u^1_j, \ldots, u^n_j \}$. Denoting $w$ to be the context window size, the objective of the auto-regressive process is to maximize the likelihood conditioned on the certain fusion of other modalities:
\begin{equation}
\begin{split}
\label{eq:autoregress}
    & \mathcal{L}_{AR} =  \frac{1}{n} \sum_{(x^i_k) \in \mathcal{D}_m} \sum_t \log P(u^t_k | f(u)),\\
    & \text{where } f(u)= f_\phi(u^{t-w}_k, \ldots, u^{t-1}_k, u^{t-w}_j, \ldots, u^{t-1}_j).
\end{split}
\end{equation}

SimVLM \cite{Wang2021SimVLMSV} improves auto-regressive (AR) pretraining via PrefixLM objective by enabling bidirectional attention on the prefix sequence, and only conducting AR factorization on the remaining tokens. It effectively utilizes the cross-modal information while reducing the training complexity.

There are also methods that conduct both auto-encoding and auto-regressive reconstruction for pretraining. For example, OPT \cite{Liu2021OPTOP} models audio, vision, and language at different granularities. At the token level, the model is trained in an auto-encoding way while at the modality level, the model performs auto-regression using modality-specific decoders to improve its generation ability. UNIMO \cite{li2020unimo} adopts both masked language modeling and seq2seq auto-regressive masked prediction.

\keypoint{Discussion}
Masked prediction has been successful in unimodal areas such as language (BERT \cite{devlin2018bert}, GPT \cite{radford2018improving}), and vision (MAE \cite{He2021MaskedAA}), and its popularity has been increasing in multimodal areas due to its ability to unify different modalities. The same masked prediction objective can conveniently be applied to all modalities after tokenizing the raw input into tokens, making it easy to scale to more modalities. 
Meanwhile, mixing different modality tokens as conditions for masked prediction can further enhance cross-modal interactions. However,
a general disadvantage of masked prediction approaches computational expense as they require extra decoders to reconstruct the input. 

Auto-encoding based masked prediction objectives are comparatively more widely adopted for multimodal learning than auto-regressive based methods. One reason for this is that AE-based objectives are faster to train than AR-based methods, because AR-based methods generate output one at a time, which is especially important when pretraining on large-scale datasets. Also, AE-based methods can better utilize bidirectional information to enhance cross-modal interactions. Nonetheless, AR-based methods have the advantage of enhancing generation ability, which is beneficial for generative downstream tasks such as image synthesis or captioning. %

\subsection{Hybrid} \label{sec:hybrid}

While single objectives already achieve good performance, many methods utilize a combination of the above approaches to take advantage of complementary strengths. This can be seen as a multi-task learning problem with several pretext tasks. A hybrid objective consisting of $N$ separate objectives can then be formulated as the weighted sum:
\begin{equation}
\label{eq:hybrid}
    \mathcal{L}_{hybrid} =  \sum_{i=1}^{N} \lambda_i \mathcal{L}_{single_i},
\end{equation}
where $\lambda_i$ is the weighting factor.%

The combination of contrastive and clustering objectives can be beneficial. As mentioned earlier, contrastive objectives may suffer from false negatives by ignoring semantic similarity between samples. On the other hand, the clustering objective takes semantic similarity into account by grouping semantically similar samples into the same cluster. MCN \cite{Chen2021MultimodalCN} performs clustering on the joint multimodal representation space (in contrast to XDC \cite{Alwassel2019SelfSupervisedLB} that clusters on the separate representation space), and also calculates a contrastive loss pairwise across audio, video, and text. The resulting high-quality embedding space enables effective retrieval of samples even from unseen datasets and domains. Furthermore, inspired by the success of contrastive learning in sound source localization (e.g., \cite{Arandjelovi2017ObjectsTS, Zhao2018TheSO}), and clustering objectives in identifying classes, Afouras et al.~\cite{Afouras2021SelfsupervisedOD} combines both objectives to learn an object detector using pseudo-labels from audio-video heatmaps and cluster labels. 

Researchers have explored hybrid objectives for vision-language pretraining, especially combining instance discrimination and masked prediction. %
For example, UNITER~\cite{Chen2019UNITERUI} employs both masked prediction and matching prediction learning at both the instance and object levels.
Contrastive learning is also widely used together with matching prediction, where the matching prediction can utilize the hard negatives calculated from the contrastive objective, enabling more grounded representation learning. 
ALBEF~\cite{Li2021AlignBF} uses contrastive learning before fusing the image and text representations and using the fused representation for MLM and matching prediction. FLAVA~\cite{Singh2021FLAVAAF} includes a similar combination of objectives, but also employs intra-modal masked modeling in order to handle the separately unpaired data. VLMO~\cite{Wang2021VLMoUV} adopts this objective as well, but employs a mixture-of-modality-experts (MOME) transformer to encode inputs with modality-specific experts.
BLIP~\cite{Li2022BLIPBL} adopts both contrastive learning and matching prediction and conducts autoregressive masked prediction to enhance generation ability.

In the field of video and language pretraining, hybrid objectives have been explored as well. ActBERT \cite{Zhu2020ActBERTLG} performs masked prediction in global actions, local regional objects, and text description levels.
UniVL \cite{Luo2020UniViLMAU} applies both masked language modeling and masked frame modeling in an auto-encoding way and language reconstruction in an auto-regressive way. It also applies contrastive to align text and video representation. MERLOT Reserve~\cite{Zellers2022MERLOTRN} outlines a novel contrastive span objective: given a video with all modalities temporally aligned, a region of text and audio are masked out. The model must maximize the similarity of the predicted masked region only to an independent encoding of the text and audio at the specific time point (positives).

Hybrid objectives are also gaining increasing popularity for video-audio pretraining. CAV-MAE \cite{gong2023contrastive} performs contrastive learning of video-audio correspondence and cross-modal masked data modeling. MAViL~\cite{Huang2022MAViLMA} proposes three multimodal objectives: (1) masked audio-video modeling; (2) masked inter-/intra-modal contrastive learning; and (3) masked self-training on multimodal features.

\keypoint{Discussion}
Hybrid objectives aim to combine complementary individual paradigms. During training, different objectives can interact, for example, contrastive learning can enhance negative pair selection for matching prediction. Furthermore, different objectives may benefit different downstream tasks, and combining them can lead to more flexible general-purpose representations \cite{Li2022BLIPBL,Li2021AlignBF}.
However, using a hybrid objective complicates hyperparameter tuning due to varying importance and differing convergence rates of individual objectives. And there is a potential risk of objective interference, where optimizing one might undermine another. In addition, such an approach could slow down training as it necessitates multiple forward passes to compute different objectives.

\section{Modality Fusion}\label{sec:architecture}
Modality fusion is the process of integrating information from diverse modalities into a unified representation, that accounts for the relationship between them and supports multi-modal tasks. Two distinct methodologies enable this: multimodal joint pretraining with various fusion stages; and ``stitching'', the amalgamation of independently pretrained unimodal models. Various architectural designs exist to conduct multimodal fusion in different stages during pretraining. Stitching, on the other hand, entails the interlinking of separately pretrained, frozen unimodal models, constructing a model capable of multimodal perception. ``Stitching'' stands out in SSML as it utilizes pretraining unimodal models using self-supervision, followed by a self-supervised fusion of multi-modalities.
We specifically focus on fusion strategies that are unique to or more popularly used in SSML.

\subsection{Fusion during Pretraining}
This subsection explores the architectures of SSML models by examining the encoder, fusion, and decoder modules. We denote the whole model as $h(x)$, and discuss each of the major architectural families below. 

\begin{figure*}[t]
\centering
\includegraphics[width=\textwidth]{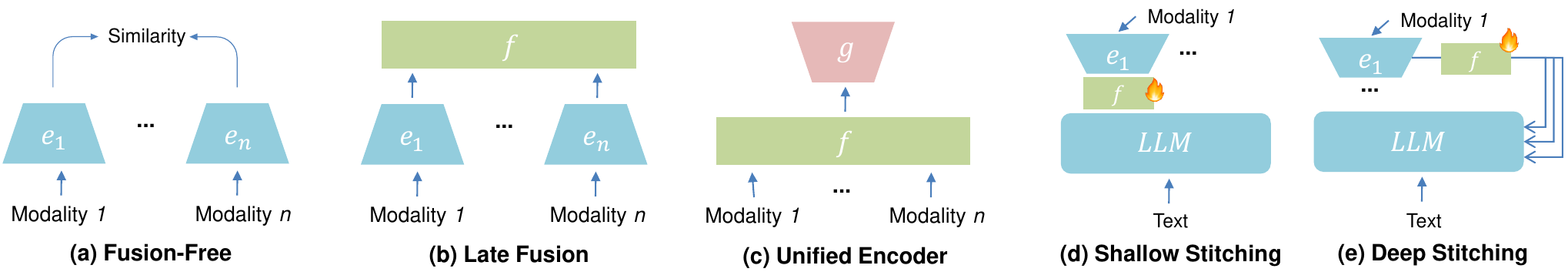}

\caption{Illustration of different modality fusion architectures.}
\label{fig:fusion}
\end{figure*}

\subsubsection{Encoders and Fusion}

\subsubsubsection{Modality-specific Encoder}
These methods adopt modality-specific encoders to encode each type of input, e.g., a CNN for visual, and a transformer for text. Then, the encoded representations can be used for simple dot product alignment or further late fusion. 

\keypoint{Fusion-free} Fusion-free methods are models that do not include an explicit fusion module $f_\psi$. Instead, they achieve cross-modal alignment by calculating similarity, such as through the use of a contrastive loss, i.e., 
\begin{equation}
    h(x) = g(e_{\theta_1}(x_1), \ldots, e_{\theta_k}(x_k)).
\end{equation}
For example, CLIP \cite{radford2021learning} and ALIGN \cite{jia2021scaling} use separate image and text encoders to pretrain on large-scale datasets with contrastive learning. MMV \cite{alayrac2020self} encodes visual, audio, and text input with three different encoders and then applies a multimodal contrastive loss for pretraining. This architecture can also be trained using matching prediction, where AVE-Net \cite{Arandjelovi2017ObjectsTS} predicts the alignment of vision and audio based on the Euclidean distance of the separate embeddings. Additionally, in XDC \cite{Alwassel2019SelfSupervisedLB}, the cluster assignments of each encoder's embedding are utilized as the supervision signal for the other encoder without further fusion. 

\keypoint{Late Fusion} Late fusion refers to models that use modality-specific encoders followed by an explicit fusion module to model the cross-modal interactions, typically via transformer layers or simply fully connected (FC) layers: 
\begin{equation}
    h(x) = g_\phi(f_\psi(e_{\theta_1}(x_1), \ldots, e_{\theta_k}(x_k))).
\end{equation}
Note that this refers to representation fusion, which is different from the traditional definition of fusing prediction probabilities from different models. For instance, embeddings from different encoders can be projected to a shared latent space and then the joint representation can be used to generate cluster assignments~\cite{Chen2021MultimodalCN} or to align predictions after a few FC layers~\cite{Owens2018AudioVisualSA}.
Using transformer layers can potentially achieve deeper modality interactions due to the use of attention mechanisms. For example, ALBEF \cite{Li2021AlignBF} and FLAVA \cite{Singh2021FLAVAAF} use an additional multimodal transformer that takes both visual and textual representations as input and fused with cross-attention, demonstrating that cross-attention benefits downstream tasks such as visual grounding/entailment/reasoning that requires in-depth modality interactions. The fusion layers can also take other forms. For instance, Dragon \cite{yasunaga2022deep} features a fusion layer with a modality interaction module that uses FC layers for information exchange. The fusion layer still retains a language encoder and a graph neural network for encoding text and the knowledge graph, respectively.

\keypoint{Discussion} 
The fusion-free architecture is popular for cross-modal alignment, as it is simple and easy to scale up. It has demonstrated strong performance on downstream tasks such as efficient retrieval. However, the lack of multimodal fusion means that modality interactions are not fully modeled, leading to poorer performance on tasks such as visual reasoning, VQA, or discrete grounding that require more complex cross-modal understanding. This issue can be addressed by incorporating a late fusion module that explicitly encourages interaction between different modalities, such as through cross-attention. However, fusion modules introduce additional computation that may slow down training compared to fusion-free models.

\subsubsubsection{Unified Encoder with Early Fusion}
An encoder capable of processing multiple modalities is referred to as a unified encoder. Such a model typically employs a transformer-based architecture to handle input tokens from various modalities, allowing it to process different types of data using a single set of parameters:
\begin{equation}
    h(x) = g_\phi((e_{\theta}(x_1, \ldots, x_k))).
\end{equation}
Here, we omit the notation of the fusion module $f_\psi$ because the fusion occurs across the entire encoder. Similar to the definition of late fusion, early fusion refers to the fusion of representations from the early stages of the unified encoder, although modality-specific tokenizers may still be required.

The tokenizer can be either an external model pretrained off-line or a model trained end-to-end jointly. External models were widely used in earlier works to extract detailed features from raw modalities which are then transformed into input token sequences. For example, object detectors have been used for image features \cite{Chen2019UNITERUI, Li2020OscarOA} and for video features \cite{Zhu2020ActBERTLG}. We refer readers to Section \ref{sec:alignment-fg} for detailed discussions. End-to-end tokenizers are trained together, and their outputs are the input tokens of the unified encoder. This can be a patch embedding, CNN, or ViT \cite{Wang2021SimVLMSV, Kim2021ViLTVT} for visual input, and MLP or transformer for text \cite{Bao2022VLBEiTGV, akbari2021vatt}.  End-to-end models have become more widely adopted as they do not require additional offline models, simplifying the training process and often yielding better performance.

Unified architectures can implicitly learn modality interactions through shared self-attention, which takes the input tokens from different modalities as input. The input token from different modalities can be distinguished, if needed, by designing specific positional encodings or extra modality type embeddings. This design is flexible, allowing for unimodal input, where only unimodal data is available and thus readily obtainable unimodal datasets can be utilized, as demonstrated in works such as \cite{Wang2021SimVLMSV, Sun2019VideoBERTAJ}. There have also been attempts to design a unified backbone for video-audio \cite{Gong2022UAVMTU}, and video-audio-text \cite{akbari2021vatt}. This allows for the mixing of tokens from different modalities, such as replacing certain text tokens with corresponding image object tokens, e.g., \cite{wang2022vlmixer}. HiP \cite{carreira2022hierarchical} scales the Perceiver-type models \cite{jaegle2021perceiver} to high-resolution raw multimodal input by building back locality into the architecture while preserving its modality-independence.
A variation on the unified architecture is the mixture-of-experts, which has been utilized recently for different modalities including image, text, video, audio, source code, etc. \cite{Wang2021VLMoUV, Wang2022ImageAA, dai2022one, shvetsova2022everything}. This design utilizes different experts for different modalities within the same multiway transformer that shares self-attention. It enables specific modality experts to be used for more precise processing, while still benefiting from the unified architecture. %

\keypoint{Discussion} 
Generally, unified architectures allow for parameter sharing among different modalities for the major components of the model, which reduces the number of parameters. Also, they are usually more robust to missing modalities compared to modality-specific encoders.
However, a major drawback of these methods is their inefficiency when it comes to tasks such as retrieval, since they require encoding all possible cross-modal pairs in order to compute similarity scores for ranking, though they can achieve competitive performance (e.g.,~\cite{Bao2022VLBEiTGV}).

\subsubsection{Decoders}
While many methods are decoder-free, others require a decoder during the pretraining phase depending on the nature of the pretext tasks. Here, the pretext head $g_\phi$ acts as the decoder. The decoder can then be discarded or retained for other downstream tasks. For example, auto-regressive masked prediction, as used in \cite{Liu2021OPTOP, Wang2021SimVLMSV, Xu2021VLMTV, Luo2020UniViLMAU, Lu2022UnifiedIOAU, seo2022end}, requires decoders to reconstruct the masked input. Successful generation, conditioned on the multimodal information of the reconstructed input, enforces the fusion of different modalities.

\keypoint{Discussion}
The additional decoder can enhance the generation ability of the model, benefiting downstream tasks such as image/video captioning and open-ended question answering.
With the popularity of large language models, many of which utilize transformer decoder-only architectures~\cite{vaswani2017attention, radford2018improving}, decoder architectures are simultaneously finding increased acceptance in multimodal learning. This brings more possibility to pretext design and flexibility for multimodal fusion.  
However, the inclusion of a decoder can also make training more computationally expensive and potentially less stable compared to decoder-free methods that only use MLPs as the output layer (e.g., \cite{radford2021learning, Arandjelovi2017LookLA, Li2021AlignBF}).

\subsection{Fusion by Stitching}
Unimodal foundation models \cite{bommasani2021opportunities}, pretrained via self-supervised learning strategies such as GPT \cite{radford2018improving}, have become pivotal in modern research due to their substantial capabilities and widespread availability. Their success, however, has given rise to a compelling challenge: How can we harness the power of these unimodal models to build multimodal models? The concept of ``stitching'', which initially referred to inserting additional layers in CNNs \cite{lenc2015understanding}, has been explored as a solution to this challenge. In the multimodal context, stitching refers to the process of integrating separately pretrained unimodal models while keeping them frozen. The connection between these models is established through an additional trainable ``stitching'' module. This module performs the task of mediating the interplay between the modalities, transforming unimodal perception into a cohesive multimodal understanding with self-supervision. This approach leverages the strengths of existing unimodal models and offers an efficient pathway towards multimodal comprehension. There are two main approaches for stitching: shallow stitching at the input layer and stitching with deep interactions.

\subsubsection{Shallow Stitching at the Input Layer}
Shallow stitching takes place at the early stage of fusion. The output representation from a pretrained unimodal model of one modality is repurposed as input for another unimodal model of a different modality, with translation usually achieved by a learned projection network.

Frozen~\cite{tsimpoukelli2021multimodal} is one of the first attempts to ground the visual modality to linguistic contexts. By treating the visual representation as soft prompts, it effectively leverages the semantic knowledge of a pretrained language model, showcasing the potential of transferring the language-only abilities to multimodal tasks in a zero-shot manner. A step further, LiMBeR~\cite{merullo2022linearly} shows that visual semantic representations can be mapped to a language space as soft prompts simply by linear layers, with both models remaining frozen. FROMAGe~\cite{koh2023grounding} demonstrates that the model can learn strong few-shot multimodal abilities with linear projections even with small-scale training data. BLIP-2~\cite{li2023blip} proposes to bridge the modality gap with a Querying Transformer using the same objectives as BLIP~\cite{Li2022BLIPBL}. It achieves state-of-the-art performance by bootstrapping the pretrained image model and language model in two stages. To alleviate the reliance on paired training data, ESPER~\cite{yu2023fusing} aligns unpaired multimodal inputs to language model generations utilizing CLIP to obtain reward signals for reinforcement learning with all models frozen. These advancements reveal a promising trajectory for lightweight and efficient multimodal learning.

\subsubsection{Stitching with Deep Interactions}
Researchers have also explored stitching models with deeper interactions. Instead of stitching only at the input layer, deep stitching fuses inside frozen models. This is typically achieved by incorporating additional adapters, and it encourages deep fusion with rich interactions.
CLIPCap \cite{mokady2021clipcap} proposes to use a transformer to map the visual tokens as the prefix embeddings for the language model, and append the prefix at each layer. Similar to Frozen, MAGMA \cite{eichenberg2022magma} introduces additional adapter modules in the form of scaled residual bottleneck MLPs between each block of the language model to better interpret visual tokens. Flamingo \cite{alayrac2022flamingo} stitches frozen vision and language models with a perceiver-based \cite{jaegle2021perceiver} module and interleaves the visual tokens with the language model across each layer with cross-attention. Remarkably, without using any purposely-annotated data, it achieves strong performance and even outperforms the fine-tuned state of the art in some tasks. However, it still requires billion-scale trainable parameters.

\keypoint{Discussion}
Stitching offers a compelling solution to multimodal fusion, thanks to its efficiency and simplicity. By harnessing the power of pretrained unimodal models, stitching significantly curtails computational and memory overhead and streamlines training. 
The simplicity is further underpinned by the common use of an auto-regressive language modeling objective, leading to strong performance with straightforward training procedures. Stitching can also leverage and retain the advanced capabilities of large-scale pretrained unimodal models, such as in-context learning~\cite{brown2020language} and chain of thought~\cite{wei2022chain}, which can be inherited without risk of forgetting due to fine-tuning.

Deep stitching introduces a relatively rich and intricate interaction between modalities by integrating the fusion inside the frozen models. However, it typically involves more complex training procedures and requires additional computational resources. The additional adapters introduced for deep stitching also make the process potentially less transparent and harder to interpret. On the other hand, shallow stitching, with its inherent simplicity, allows for efficient implementation and straightforward training, promoting the effective transfer of abilities to multimodal tasks. Nevertheless, the fusion of modalities is relatively superficial, potentially limiting the degree of fusion between modalities. How to bootstrap knowledge from pretrained unimodal models and fuse the frozen modality-specific representations remain a promising avenue for future research.

Given its progress in the vision and language domains, extending stitching to other modalities presents an exciting opportunity for future research. However, there is a risk of unwittingly inheriting biases embedded in pretrained unimodal models. This calls for scrutiny and active bias mitigation strategies during stitching. Furthermore, we need a better understanding of the distinct properties and possible interactions of the frozen unimodal representations. Analyzing the differences and similarities in representations of different modalities can improve the interpretability and transparency of stitched multimodal models.

\section{Unaligned Multimodal Learning}\label{sec:data}

\begin{figure}[t]
\centering
\includegraphics[width=\linewidth]{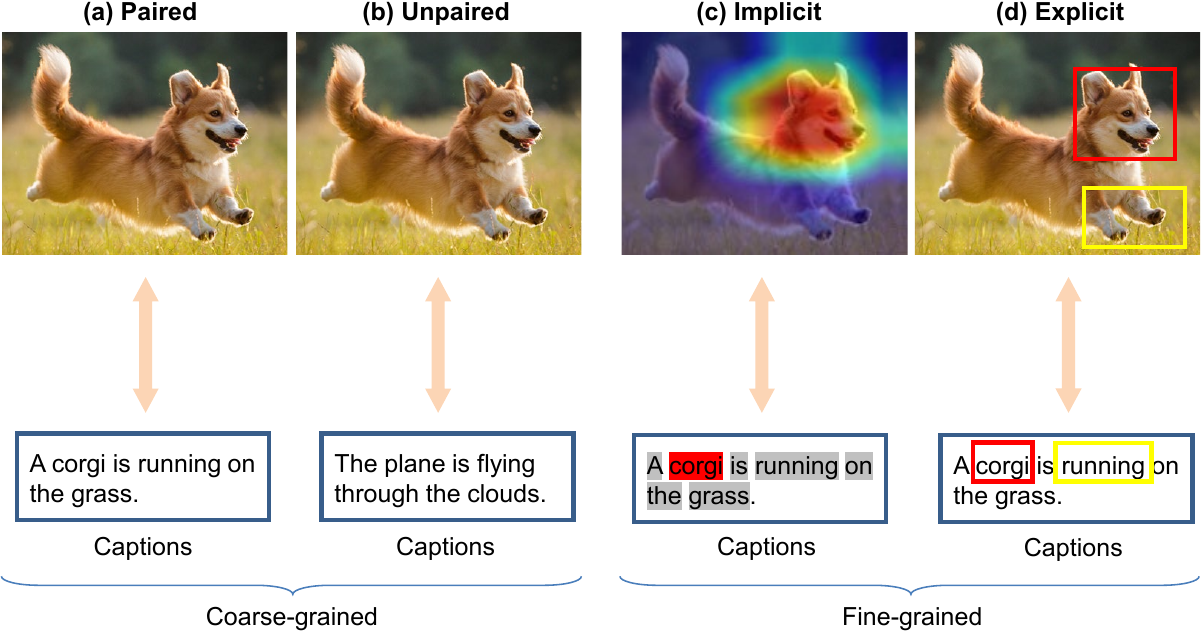}

\caption{Illustration of (a) coarse-grained paired input, (b) coarse-grained unpaired input, (c) fine-grained implicit alignment, and (d) fine-grained explicit alignment.}
\label{fig:alignment}
\end{figure}

The challenge of unaligned multimodal data is a unique dimension to SSML. Alignment arises at both coarse-grained and fine-grained scales (Fig.~\ref{fig:alignment}). Coarse-grained alignment pertains to instance-level pairing, that is, pairing of data samples across different modalities. In contrast, fine-grained alignment involves correspondence between sub-components of these instances, such as linking objects in an image with corresponding words in a sentence. Defining the correspondences above is often a manual annotation process, and thus self-supervised learning ideally calls for methods that do not require alignment to be annotated.

\subsection{Coarse-grained}
We consider three scenarios for coarse-grained input pairing: where all modalities are coarsely aligned (paired), where there is no correspondence known between modalities (unpaired), and where there is a mix of these two data types (mixed). We devote more attention to the latter two cases, which pose greater challenges.

\subsubsection{Noisy-Paired}
Assuming paired data, with known correspondence between samples in each modality, is the most popular setting in multimodal learning. Paired data occurs naturally in cases where the modalities are recorded synchronously (e.g., audio and video modalities in a video clip), but can also be created incidentally in other cases (e.g., alt text tags of web images). However, the data pairing can be noisy in both cases. For example, in naturally paired data like audio-video, a speaker might discuss a topic before physically demonstrating it, or they might neglect to describe a visible action. Similarly, in web-crawled image-text pairs, the text might only partially cover image content.

To learn representations from videos where noisy pairs can occur, methods have been developed to exploit temporal alignment among visual, audio, or text modalities using different strategies. For example, negative pairs can be selected from different videos \cite{Arandjelovi2017LookLA, Arandjelovi2017ObjectsTS, alayrac2020self}, from the same video but using modalities from different frames \cite{Owens2018AudioVisualSA}, or by selecting multiple correct positives while downweighting incorrect ones~\cite{miech2020end}.
Image-text pairs are a special case as captions must be manually created from images. However, as a large amount of noisy image-text pairs can be easily crawled from the internet, researchers have achieved promising results with various objectives as described in Section \ref{sec:Objective} even with noisy-paired data, e.g., \cite{radford2021learning, Li2022BLIPBL, Wang2022ImageAA, Tschannen2022ImageandLanguageUF}.

\keypoint{Discussion}
While the use of approximately paired data in multimodal learning can benefit from straightforward pretext tasks that assume known pairing, it remains to be seen if such alignment noise ultimately limits the efficacy of these models.
Furthermore, in many domains obtaining paired data is still a bottleneck. For example, in healthcare, obtaining paired data can be impossible due to privacy reasons. Even in the image-text setting, where substantial paired data exists, the volume of unpaired images and text vastly dwarfs the volume of paired examples. Thus it is desirable to develop methods capable of learning where some or all of the data is unpaired. 

\subsubsection{Unpaired}
Unpaired learning aims to reduce the dependency on well-aligned multimodal data pairs and directly leverage large-scale unimodal data corpora for multimodal learning. The key goal of unpaired multimodal algorithms is to find ways to induce alignment in unaligned data. Existing approaches mainly achieve this in two ways: (1) using external models, or (2) enforcing cycle-consistency.

External models are often used to detect concepts that enable connections between instances in one modality with similar instances in another, thus creating noisy coarse-grained pairs to provide extra supervision for multimodal learning. Object detectors, sometimes called concept detectors, are often used to extract region features and object tags to align vision and language data. For example, U-VisualBERT \cite{Li2021UnsupervisedVP}, VLMixer \cite{wang2022vlmixer}, and $\mu$-VLA~\cite{Zhou2022UnsupervisedVP} utilize object detectors to extract the object tags from the images and then connect to the text with the concept words. Other tools such as scene graph detectors have also been used to extract the mutual relationship. For example, Graph-Align \cite{Gu2019UnpairedIC} enables unsupervised image captioning by extracting both scene graphs from both text and images.
We discuss in detail the use of external models for explicitly extracting fine-grained pairing in Section~\ref{sec:alignment-fg}.

Enforcing cycle consistency \cite{zhu2017unpaired} is another approach to aligning the representation of different modalities. In the multimodal context, consider two modalities $A$ and $B$, where we want to learn a mapping $G_{AB}: A \rightarrow B$ and $G_{BA}: B \rightarrow A$. The concept of cycle consistency encodes the intuition that these mappings should be the reverse of each other and that both mappings should be bijections. Specifically, a cycle consistency loss encourages $G_{AB} \left(G_{BA}\left(b\right)\right) \approx b$, and vice versa. DM2C \cite{Jiang2019DM2CDM} leverages cycle consistency to learn the unpaired cross-modal mappings of the latent representation with a special inter-modal auto-encoder, which is first pretrained on the single-modal data. MACK \cite{huang2022mack} collects a set of conceptual words and their related image regions from publicly available datasets, and then computes prototypical region representations to obtain the pretrained general knowledge. To further fine-tune the model for specific datasets a  region-level cycle-consistency loss can be applied. Similarly, Graph-Align \cite{Gu2019UnpairedIC} also adopts this loss for modality alignment.

\keypoint{Discussion} 
Learning with unpaired multimodal data is challenging but of great practical value as it enables the use of large-scale unimodal data corpora. Although this has gained growing attention in recent years, it is still immature. Approaches that rely on external pre-trained models are limited by the external model's quality, and these may not generalize well to the specific task or dataset being used. Furthermore, external models limit generality and scalability, as pretrained models covering all concepts or domains of interest may not exist. 
Meanwhile, cycle-consistency losses typically require additional model components, e.g., decoders. This adds additional complexity to the model and increases the amount of computation required. %

\subsubsection{Mixed}
Training with a mix of paired and unpaired multimodal data is often the most realistic scenario as it reflects the reality that there is often some paired data and a larger amount of unpaired data available. Methods for dealing with mixed-pairing data usually apply separate pretext tasks for unpaired data and paired data pretext tasks for paired inputs in a multi-task manner.

Masked prediction is widely used for dealing with mixed-paired data due to its flexibility. VATLM \cite{Zhu2022VATLMVP} uses a unified masked prediction objective, with within-modality masked prediction for text and multimodal masked prediction from combinations of visual-audio-text pairs. 
BEiT-3 \cite{Wang2022ImageAA} pretrains a multiway transformer by performing masked data modeling on image, text, and image-text pairs.

Clustering methods have also been utilized. For example, u-HuBERT \cite{hsu2022u} adopts separate audio and video encoders, and predicts the cluster assignments of the masked input frames after they are concatenated by a fusion module. If any of the modalities are missing, a dummy concatenation (i.e., by adding zero) is performed, and the same objective function can then be used as usual.

In terms of using different objectives for paired and unpaired data, VideoBERT \cite{Sun2019VideoBERTAJ} features a unified architecture and performs masked prediction for both text-only and video-only data. When paired data is available, it utilizes matching prediction to learn cross-modal correspondence.
FLAVA \cite{Singh2021FLAVAAF} employs masked image modeling and masked language modeling for image-only and text-only data, while it utilizes masked multimodal modeling and contrastive learning over paired data. UNIMO \cite{li2020unimo} applies masked image modeling for image-only data, both AE-based and AR-based masked prediction for text-only data.
It utilizes both unimodal and multimodal data, retrieving similar unimodal samples as positive pairs for cross-modal contrastive learning.
VLMO \cite{Wang2021VLMoUV} adopts a stagewise training approach that starts by training image-only and text-only modality experts using masked prediction, followed by instance discrimination on image-text pairs. Notably, SkillNet \cite{dai2022one} utilizes mixture-of-experts for five different modalities, and can be trained on paired image-text and video-text using contrastive learning, and on unpaired sound and code using clustering and masked prediction.

\keypoint{Discussion}
Methods of mixed-pair learning can effectively leverage readily available unimodal datasets, leading to improved scalability and thus better downstream performance. However, some limitations do exist. For example, if the volume of unimodal data is imbalanced, the trained models may suffer from imbalanced performance on different downstream tasks or can overfit certain modalities.

\subsection{Fine-grained}
The previous subsection covered the limited coarse-grained pairing as supervision. In this subsection, we discuss techniques to induce fine-grained multimodal alignment (Fig.~\ref{fig:alignment}~(c,d)) from coarse supervision (Fig.~\ref{fig:alignment}~(a)). %
We distinguish between explicit alignment methods which infer a discrete grounding or correspondence between sub-elements or tokens within each modality (Fig.~\ref{fig:alignment}~(d)), and implicit alignment methods which infer a soft association between tokens in each modality (Fig.~\ref{fig:alignment}~(c)).

Here, we present a general formalization for fine-grained alignment.
For each instance, suppose we have a set of fine-grained elements (embeddings/tokens/patches) $\Omega_k = \{\omega_k^1, \ldots, \omega_k^l \}$ from each modality $x_k$, where $l$ is the number of the elements. For clarity, we assume the number of elements is the same for different modalities.
For modality $x_i$ and $x_j$, the fine-grained alignment tries to find a permutation function $\pi$ over $\{1, \ldots, l\}$, such that the elements in $\Omega_i$ are aligned with the embeddings in $\Omega_j$. We can further write the corresponding permutation matrix as $\Pi \in [0, 1]^{l \times l}$. Implicit and explicit alignment corresponds to constraints imposed on $\Pi$. For implicit alignment, we have $\Pi 1_n = 1_n$ and $\Pi^\top 1_n = 1_n$, where $1_n$ is an $n$-dimensional vector with all ones.
For explicit alignment with multi-instance learning, we enforce the discrete correspondence that $\Pi \in \{0, 1\}^{l \times l} \text{and } \Pi 1_n = 1_n$. $\Pi_{pq}$ is set to 1 if token $q$ in modality $x_i$ corresponds to token $q$ in modality $x_j$, and 0 otherwise (the values are continuous for implicit and discrete for explicit). When the number of elements varies across modalities, the constraint of the permutation matrix will be relaxed. Most fine-grained alignment methods correspond to optimizing a loss of the form:
\begin{equation}
\mathcal{L}_{fg} = \Gamma(\Omega_i, \Pi \Omega_j),
\end{equation}
with respect to $\Pi$ and the representation parameters that produce $\omega$, and where $\Gamma(\cdot,\cdot)$ is a function to measure distance, such as L2-norm. I.E: Simultaneous correspondence and representation learning.

\subsubsection{Implicit}
Implicit alignment is often achieved by enforcing cross-modal connections in the embedding space. Cross-modal attention and optimal transport are two commonly used techniques (i.e., realizations of the permutation function) for achieving such connections.

Self-attention is a powerful mechanism that allows elements of input set to interact \cite{vaswani2017attention}. In multimodal learning, self-attention is extended to cross-attention, and the inferred attention map induces fine-grained correspondence between the modalities.
LLA-CMA's \cite{Cheng2020LookLA} co-attention module consists of audio-guided attention and visual-guided attention, allowing the model to exploit audio-visual co-occurrence. ViLBERT \cite{Lu2019ViLBERTPT} introduces a co-attentional layer to produce attention-pooled features for both image and text modalities, thus enabling sparse interactions between them. Similarly, co-attention is also used in FLAVA \cite{Singh2021FLAVAAF} and SelfDoc \cite{Li2021SelfDocSD} to uncover inter-modal relationships. %

Cross-attention can also model global-local interactions. For instance, ActBERT \cite{Zhu2020ActBERTLG} incorporates additional global-local correspondences into its design by stacking original key-value pairs with values from the other modality, thus ensuring that the joint video-text representation is aware of both fine-grained objects and global information. COOT~\cite{ging2020coot} optimizes representations with respect to interactions between local features (clips/words) and global context (frames/sentences) by inputting local representations as key-value pairs and the global representation as the query.

Cross-modal attention can also be conducted in a directed manner, where one modality attends to another, but not vice versa. This approach is designed with the idea that some modalities require more complex modeling, while other modalities can be adequately encoded with a shallower model.
For instance, ALBEF \cite{Li2021AlignBF} fuses the image representation into the multimodal encoder to align the unimodal text representation. Similarly, BLIP~\cite{Li2022BLIPBL} uses an image-grounded text encoder and decoder to fuse the visual representation with texts through cross-attention.

{Recent methods such as \cite{Wang2022ImageAA, Wang2021VLMoUV} employ shared} self-attention for different modalities encoded by a mixture of experts. Although not explicitly studied, the shared self-attention weights have the potential to establish connections between different modalities. To demonstrate the fine-grained alignment between modalities, several methods such as Oscar \cite{Li2020OscarOA}, UNITER \cite{Chen2019UNITERUI}, Hero \cite{Li2020HeroHE}, and $\mu$-VLA \cite{Zhou2022UnsupervisedVP} have used visualized the learned attention weights. They show that attention can learn cross-modal alignment, such as mapping words to image regions.

Optimal transport (OT) \cite{peyre2019computational, chen2020graphOT}, which defines distances between probability measures, is also used for the cross-domain fine-grained alignment. OT for cross-domain alignment aims to match the distributions by minimizing the cost of transforming one distribution into another. UNITER \cite{Chen2019UNITERUI} employs OT to minimize the cost of transporting representations from image regions to words in a sentence (and vice versa), resulting in improved cross-modal alignment. The fast inexact proximal point method for optimal transports (IPOT) \cite{xie2020fast} is used to approximate the OT distance to overcome the challenge of intractable computation. Similarly, ViLT \cite{Kim2021ViLTVT} adopts this approach to align textual subsets and visual subsets, which are not extracted by external models, unlike in UNITER \cite{Chen2019UNITERUI}.

Canonical correlation analysis (CCA) \cite{hardoon2004canonical} is a classic approach to finding linear relationships between two sets of variables, while ensuring orthogonality. Its objective is to maximize the correlation between the corresponding dimensions of the representations from different modalities. While deep extensions of CCA exist \cite{andrew2013deep,chang2018scalable}, and have been applied to tasks such as fine-grained audio-visual correlation \cite{passos2023multimodal}, it has not been widely used in SSML overall.

\subsubsection{Explicit}
In contrast to implicit alignment, methods have also been developed to introduce explicit alignment. Explicit fine-grained pairing refers to the correspondence between smaller, more specific components of the instances, such as objects in an image with words in a sentence. This type of correspondence can be introduced by external models or via multi-instance learning.

\subsubsubsection{Multi-instance Learning}
In contrast to implicit correspondence, explicit alignment-based methods usually use fine-grained elements $\omega$ that are produced as a result of a different process such as saliency detection. Multiple Instance Learning (MIL) \cite{dietterich1997solving} is a commonly used approach for explicit alignment where there is a one-to-many correspondence between elements across modalities. To localize the sound source, AVOL-Net \cite{Arandjelovi2017ObjectsTS} extracts local region-level image descriptors on a spatial grid and then computes a similarity score between the audio embedding and each of the vision descriptors. The maximal similarity score is used as the measure of the image-audio agreement, i.e., correspondence score, to train the network. This approach encourages one image region to respond highly to the corresponding audio and therefore localize the object. DMC \cite{Hu2018DeepMC} proposes to extract audiovisual entities by aggregating the similar feature vectors between the audio and visual modalities based on the assumption that the elements in the feature maps have similar activation probabilities for the same unimodal component. Thus, it clusters the unimodal feature vectors into object-level representations, and aligns them in the audiovisual environment.

Explicit alignment methods are less commonly studied than their implicit correspondence counterparts, but benefit from greater interpretability of the learned correspondence. However, they are sensitive to the process used to generate the set $\Omega_k$ for correspondence and the validity of assumptions correspondences (one-to-one, one-to-many, etc.). 

\subsubsubsection{External Models}
\label{sec:alignment-fg}
Explicit fine-grained input pairing can help the model learn detailed relationships between different modalities. In this survey, we define SSML methods as those that do not require manual annotations. Thus these methods may not strictly be considered self-supervised as the external models may be trained on datasets using supervision. However, they are included due to their ease of access through open-source communities and for completeness.

For visual data, object detectors (e.g., Faster R-CNN \cite{Ren2015FasterRT}) are frequently used to extract regions of interest (ROI) and the object classes. They can then be used to align with the corresponding parts in the other modalities. For image-text pretraining, extracted ROIs can be used for word-region alignment \cite{Chen2019UNITERUI, li2020unimo, wang2022vlmixer, Lu2019ViLBERTPT, Su2019VLBERTPO}, masked object classification \cite{Su2019VLBERTPO, Li2019UnicoderVLAU}, and feature regression \cite{Tan2019LXMERTLC}. It can also be used to extract ROI from videos in every static frame at a set framerate, e.g., ActBERT \cite{Zhu2020ActBERTLG}. For document understanding, SelfDoc \cite{Li2021SelfDocSD} extracts document object proposals and applies OCR to obtain words for each proposal.

Object detectors can also be used to align unpaired data. To achieve noisy alignment, U-VisualBERT \cite{Li2021UnsupervisedVP} uses a pretrained detector to extract the object tags from the images, which are appended to the token embeddings with spatial coordinates. A masked prediction objective is then also applied to the detected tags to provide a noisy grounding signal via reconstruction. VLMixer \cite{wang2022vlmixer} proposes to randomly wipe off some concept words in the sentence and then paste the visual patches with the same concept labels generated by detectors to obtain mixed sentences, serving as a cross-modal representation of the original sentence.
Then, masked language modeling and contrastive learning are used to learn cross-modal alignment. $\mu$-VLA~\cite{Zhou2022UnsupervisedVP} weakly aligns an image-text corpus with a retrieval-based method and employs multi-granular alignment pretext tasks including masked prediction, contrastive learning, and the classification of the detected object labels.

To extract various levels of features from language, pretrained semantic parsers are used to obtain an explicitly factorized semantic space. 
For example, UNIMO \cite{li2020unimo} applies a scene graph parser to collate objects, attributes, and relations into vocabularies, which then aid in data augmentation through text rewriting at various levels.
The extracted scene graph can also be applied to tasks such as unpaired image captioning \cite{Gu2019UnpairedIC} by aligning language and image information through parsing sentences syntactically.

\section{Theoretical Considerations}

Current research on SSML is predominantly empirical, with limited theoretical analysis. Notably, many of the prominent theoretical frameworks developed for unimodal self-supervised learning~\cite{saunshi2019theoretical} do not directly apply to the multimodal context. One of the few theoretical frameworks that has a known extension to SSML is the information bottleneck principle (IB)~\cite{tishby2000information}. Classic IB interprets supervised learning in information theoretic terms, as a process of learning a representation that maximizes MI between the encoding and the label  $Y$, while minimizing information between the input $X$ and the hidden representation $R$:

\begin{equation}
    L = I(X; R) - \beta I(R; Y)
\end{equation}
where $I(A; B)$ denotes the mutual information between $A$ and $B$, and $\beta$ is a trade-off parameter controlling the balance between compression and relevance. In SSML, the objective extends to preserving relevant shared information across modalities while compressing modality-specific irrelevant information. Denote $X_i$ to be the $i$-th modality, $R_i$ is the latent representation of $X_i$, and $Z$ to be the pseudo target.
The IB objective for multimodal data can be expressed as:

\begin{equation}
    L = \sum_{i=1}^m \alpha_i I(X_i; R_i) - \gamma I(R_1, R_2, \ldots, R_m; Z)
\end{equation}
where $I(X_i; R_i)$ quantifies the information retained about modality $X_i$ in $R_i$, \(I(R_1, r_2 \ldots, R_m; Z)\) measures the joint information in \(R_1, R_2, \ldots, R_m\) about the $Z$, and $\alpha_i$ and $\gamma$ are trade-off parameters. 

Different from unimodal multi-view SSL~\cite{shwartz2024compress}, substantial modality-specific information exists in SSML, violating the assumption that all relevant information is shared across views. This discrepancy necessitates adapting the IB framework to handle both shared and unique information:
\begin{equation}
    L = \sum_{i=1}^m \alpha_i I(X_i; R_i) - \sum_{i \neq j} \beta_{ij} I(R_i; R_j) - \gamma I(R_1, \ldots, R_m; Z),
\end{equation}
where $\beta_{ij}$ are trade-off parameters controlling the balance between modality-specific and shared information. $I(R_i; R_j)$ measures the mutual information between the representations of different modalities. This objective encourages the model to capture the essential shared information while preserving modality-specific details, thus improving the generalizability of the learned representations.

However, the overall theoretical analysis of SSML remains an outstanding challenge, presenting significant opportunities for future research to develop comprehensive theoretical frameworks that can adequately address the complexities of multimodal data.

\section{Applications}\label{sec:applications}
SSML algorithms have been widely applied to real-world scenarios, including state representation learning, healthcare, remote sensing, and many other fields, such as autonomous driving \cite{Valverde2021ThereIM, Sautier2022ImagetoLidarSD} and machine translation~\cite{Dabre2019ASO}.

\subsection{State Representation Learning for Control}
State representation learning (SRL) is a special type of multimodal representation learning that captures the interaction between environmental observation modalities and an agent's action modality. SRL need not be task-specific and can be solved with SSML objectives. The learned representation can then be transferred to benefit downstream reinforcement learning and control tasks. 

A common approach to SRL is masked prediction. Forward model and inverse models \cite{lesort2018state, botteghi2022unsupervised} in control can be considered a special form of auto-regressive masked prediction. {In the classic reinforcement learning framework~\cite{sutton1998introduction}, the \textit{observation} is the raw sensor information and the \textit{state} is a compressed depiction of this information that contains the necessary information for \textit{action} selection. A forward model predicts the future state $s_{t+1}$ from current action $a_t$ and current observation or state $o_t/s_t$; while an inverse model predicts action $a_t$ given observations $o_t$ and $o_{t+1}$ or states $s_t$ and $s_{t+1}$}. The unobserved state/action can be considered masked and SRL corresponds to SSML masked prediction. For example, \cite{Oh2015ActionConditionalVP} proposes an action-conditional video forward model that predicts the next $k$ states in the future. The World Model \cite{ha2018recurrent} is trained to predict the future representation given the past visual observations and actions. Then an agent decides what actions to take based only on the learned representations to get rewards. Similarly, PlaNet \cite{hafner2019learning} learns the environment dynamics from images and chooses actions through planning in the latent space, where the dynamics model, containing both deterministic and stochastic transition components, learns to predict the rewards for multiple time steps ahead. \cite{agrawal2016learning} trains a model by having a robot randomly poke objects and recording before/after visual states, estimating forward and inverse models of dynamics. Masked prediction objectives can also be used together with instance discrimination. For instance, Contrastive Forward Modeling \cite{yan2021learning} combines contrastive predictive coding \cite{Oord2018RepresentationLW} with predicting future state to maximize the mutual information between predictions and positives. \cite{Lee2018MakingSO} propose to fuse representations of images, force, and proprioception and then learn to predict the next control cycle's optical flow and potential environmental contact, using auto-encoding and matching prediction, respectively.

\subsection{Healthcare}
Clinicians often rely on information from multiple sources and modalities in diagnosis, prognosis, and treatment planning. Representation learning on diverse data is thus important for accurate diagnoses and effective patient care. Medical imaging is widely used for automatic diagnosis, and various methods have been proposed to improve downstream diagnosis with imaging and other modalities. 
For example, ConVIRT \cite{zhang2020ConVIRT}, GLoRIA~\cite{Huang2021GLoRIAAM}, and CheXzero~\cite{Tiu2022ExpertlevelDO} adopt contrastive learning for medical imaging and medical reports.
MEDCLIP \cite{Wang2022MedCLIPCL} decouples images and text, utilizing readily available image-only and text-only training data. ContIG \cite{taleb2022contig} aligns retinal images and genetic modalities with a contrastive loss. CoMIR \cite{pielawski2020comir} enables registration with a contrastive loss by enforcing rotational equivariance. %

\subsection{Remote Sensing}
In remote sensing, different sensors can provide complementary information for earth observations, including hyperspectral data, multispectral data, light detection and ranging (LiDAR), synthetic aperture radar (SAR) data, etc. \cite{Wang2022SelfsupervisedLI}. For instance, hyperspectral images capture land-cover categories through spectral signatures, while SAR images offer dielectric properties. Techniques to integrate these datasets include using contrastive loss for SAR and optical images \cite{Chen2022SelfSupervisedSD}, and masked prediction for multispectral and SAR data \cite{Montanaro2021SemiSupervisedLF}. Change detection, crucial for areas like damage assessment, employs methods such as 
clustering with contrastive learning for analyzing bi-temporal scenes from different sensors \cite{Saha2021SelfSupervisedMC}. Additionally, geo-tagged audio recordings are used to establish correspondence with image data via contrastive learning \cite{Heidler2021SelfsupervisedAR}.

\section{Challenges and Future Directions} \label{sec:discussion}

\keypoint{Scaling} 
One of the most compelling attributes of self-supervision is its scalability. The progression of large language models (LLMs)~\cite{brown2020language, touvron2023llama} has further encouraged the expansion of multimodal LLMs trained via self-supervision~\cite{yin2023survey}. Regarding objectives, the auto-regressive approach has demonstrated efficacy in training large multi-modal models by predicting the next multimodal ``word''~\cite{driess2023palme, ge2024seed, sun2024emu}. The unified objective is also easy to scale. In terms of modality fusion, the ``stitching'' technique leverages pretrained unimodal models, subsequently enhancing them with multi-modal training, efficiently scaling up larger multi-modal models with reduced resource requirements. While specific techniques remain as open challenges, stitching and auto-regression serve as popular recipes for scaling larger multimodal models.

\keypoint{Resources}
Major challenges of SSML include high computational demands and limited access to large-scale, publicly available datasets. Training models with billions of parameters requires vast amounts of data, and while efforts like LAION~\cite{schuhmann2022laion} exist, the scarcity of public datasets hinders organizations without proprietary data, risking AI de-democratization~\cite{Ahmed2020TheDO}.
Efforts to improve SSML training efficiency include masked token dropping \cite{li2022flip} and decoupled gradient accumulation \cite{cui2022contrastive}. Parameter sharing among modalities \cite{Sun2019VideoBERTAJ, Su2019VLBERTPO, Hendricks2021DecouplingTR} and attention weight sharing with a mixture-of-experts \cite{Wang2021VLMoUV, Wang2022ImageAA} also help reduce the number of parameters.
To reduce computational resources, data and model pruning techniques~\cite{Sorscher2022BeyondNS, zhang2022platon} and better utilization of existing paired data~\cite{li2022supervision} can be applied. However, balancing state-of-the-art performance with efficiency remains an open question.

\keypoint{Data Acquisition and Noise}
Given the reliance on ``free labels'' such as web-crawled data, many SSML methods are trained on noisy paired samples in practice, which is a unique challenge compared to unimodal data. Related to learning from noisy labels \cite{natarajan2013learning, song2022learning}, it is important to study how much noise can be tolerated while still learning a good representation in order to trade off training data size and quality. 
Current methods have explored scaling to a large number of noisy pairs \cite{jia2021scaling} or bootstrapping high-quality data pairs from the noisy datasets \cite{Li2022BLIPBL}. However, data filtering steps in many methods are not explicitly discussed as part of the machine learning process. We recommend that data crawling and filtering procedures should be considered as first-class algorithms for study and dissemination, just like neural architectures and objectives, since they can crucially influence the final performance. 
Finally, as the outputs of generative models including SSML methods are shared on the internet, future iterations of these models are likely to be trained on artificially generated data, which could be detrimental to performance. The impact of such feedback cycles on datasets is not yet understood. 

\keypoint{Unpaired, Mixed, and Interleaved Data}
Modeling unpaired or mixed-pairing data remains challenging due to the additional correspondence ambiguity. Some methods rely on external models to make connections between different modalities, such as using an object detector to extract object tags for alignment with sentences \cite{li2020unimo, wang2022vlmixer, Zhou2022UnsupervisedVP, Wang2022MedCLIPCL}. However, such approaches depend on the quality of the pretrained model. %
Furthermore, they rely on the assumption that there is overlapping semantic content between the unpaired modalities, which may limit the applicability of the models where each modality is independently sampled. It is crucial to quantify how robust the unpaired models are to varying degrees of underlying correspondence, while completely unconstrained/uncurated cases remain a challenging problem that has not been adequately studied in previous works. Additionally, current unpaired/mixed methods are typically trained on datasets with a relatively balanced number of samples from each modality. However, as imbalanced datasets are more common in practice, it is important to evaluate their sensitivity to imbalanced training samples and to develop new methods to address this issue. Moreover, arbitrarily interleaved data, such as web pages with images and text, provide a valuable opportunity for SSML. These naturally co-occurring multimodal examples are abundantly and easy to obtain, offering diverse contextual associations between modalities that could considerably enhance performance \cite{alayrac2022flamingo}. However, this form of data remains relatively underexplored. Future endeavors should consider creating more interleaved multimodal datasets and devising new methodologies to realize the potential.

\keypoint{Robustness and Fairness}
As SSML models become more widely used, it is crucial to ensure their reliability prior to deployment. However, state-of-the-art models are still limited in this regard. For example, vision-language models have been shown to have limited compositional understanding and struggle with simple concepts like ``behind''~\cite{yuksekgonul2023when}. %
Another significant challenge in robustness is to maintain stable performance when arbitrary combinations of modalities are added or removed at inference time. SSL has shown itself to be promising in improving robust multimodal representations~\cite{mckinzie2023robustness}.
Also, despite the growing interest in fairness issues in generative models \cite{Carlini2023ExtractingTD}, discriminative SSML models are not immune to bias. For instance, they have been found to exhibit bias based on gender \cite{srinivasan2022worst} or race~\cite{Agarwal2021EvaluatingCT} by aggregating bias from multimodal data \cite{Birhane2021MultimodalDM}, the model architecture and objective function \cite{liang2022mind}. Future work on identifying and eliminating the sources of bias is required.

\keypoint{Unification}
The trend towards unification in SSML has emerged across three principal axes: architectures, objective functions, and tasks.
The first facet, architectural unification, entails the use of a specific modality encoder to process all input modalities (e.g., a vision model to encode language \cite{rust2023language}) or a specifically designed architecture to encode multiple modalities \cite{wu2019unified, dai2022one}. This unification facilitates an adaptable architecture that can cater to various modalities with increased efficacy.
Similarly, the consolidation of pretraining objectives into a common target, as seen in approaches like data2vec~\cite{Baevski2022data2vecAG} and BEiT-v3 \cite{Wang2022ImageAA}, has enabled an easier integration of modalities. By employing a common objective across different modalities, the learning process becomes more streamlined and enables more straightforward adaptation to a wide range of data types, such as language models for proteins~\cite{madani2023large}.
Lastly, unification in SSML also refers to the presentation of various tasks in a consistent form. For example, image classification, segmentation, and detection can be unified by text generation. This harmonization of tasks allows for zero-shot performance on previously unseen tasks, markedly enhancing multimodal models' generalizability across various domains.

\keypoint{Emergent Abilities}
Large language models exhibit emergent abilities, where meaningful performance on certain tasks can only be observed when the model size reaches a certain scale~\cite{wei2022emergent}. However, emergent abilities in multimodal models remain relatively unexplored. SSL allows for the scaling of multimodal models, which may lead to emergent abilities in challenging tasks such as multimodal reasoning, bringing us closer to general-purpose models. %
Notably, emergent abilities may manifest differently between modalities. For instance, language models appear to gain common sense and logic abilities from scaling \cite{brown2020language} while a multimodal model may showcase emergent zero-shot recognition \cite{girdhar2023imagebind}. Analyzing how and when these properties emerge across modalities can provide insights into the representations learned. Additionally, emergent multimodal abilities may arise from model architectures uniquely suited for fusing modalities or the alignment mechanism. Identifying fusion and alignment techniques that unlock new abilities will underpin future multimodal model designs.

\keypoint{Rethinking Self-supervision in the Multimodal Context and Beyond}
The core appeal of SSL is utilizing widely available unlabeled data \cite{balestriero2023cookbook}. However, applying this principle to multimodal data raises open questions. First, multimodal data differs from unimodal data, for which SSL paradigms are well-established. As discussed in Sec~\ref{sec:scope}, should we consider freely available co-occurring web-crawled data pairs as fulfilling the self-supervision criteria? Second, new paradigms are emerging, especially with large language models, the outputs of which could potentially offer unlimited sources of supervision at scale. For instance, can text generated by a model like ChatGPT~\cite{openai2023chatgpt} be considered a valid form of self-supervision given the manually-designed template? 
On one hand, these outputs do require a modicum of manually-provided prompt templates. On the other hand, the design of prompt templates can be analogous to manually-defined rules for web data crawling or the design of pretext tasks (e.g., prompt template: rewriting image caption v.s.~pretext task: contrastive learning with different augmentations). 
As SSML research progresses, our understanding of what constitutes ``self-supervision'' is likely to evolve. Careful consideration of the spirit and aims of SSL will be important for developing methods that learn rich multimodal representations from abundant data.

\section*{Acknowledgments}
Yongshuo Zong is supported by the United Kingdom Research and Innovation (grant EP/S02431X/1), UKRI Centre for Doctoral Training in Biomedical AI at the University of Edinburgh, School of Informatics. For the purpose of open access, the author has applied a creative commons attribution (CC BY) licence to any author accepted manuscript version arising.

\ifCLASSOPTIONcaptionsoff
  \newpage
\fi

\bibliographystyle{IEEEtran}
\bibliography{ref}

\clearpage

\onecolumn
\appendices 
The supplementary material includes (1) an overview of multimodal downstream tasks and (2) Table~\ref{table:algorithms-contrastive}-\ref{table:algorithms-hybrid}, which present collections of methods with different objectives, as well as information on alignment and architecture. Additionally, we provide a summary of the latest advances in methods, datasets, and implementation, which can be found at \url{https://github.com/ys-zong/awesome-self-supervised-multimodal-learning}.

\section{Multimodal Downstream Tasks}

\subsection{Vision-Language Downstream Tasks}
\begin{figure*}[h]
\centering
\includegraphics[width=\textwidth]{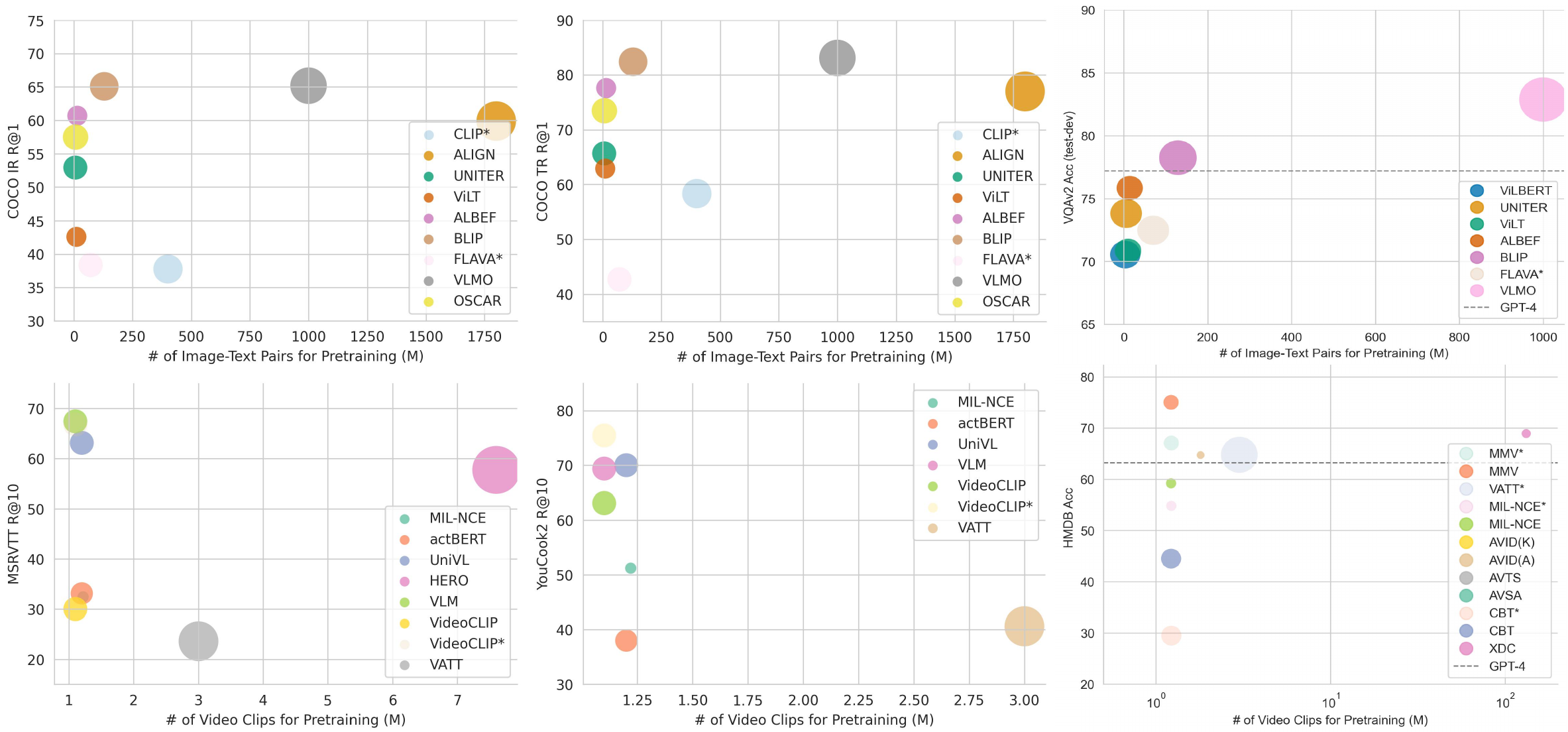}

\caption{Results of image-text model and multimodal video models on downstream tasks. The size of each scatter corresponds to the number of estimated model parameters. \textbf{Top row}, recall@1 of text-to-image and image-to-text retrieval on COCO dataset, and the accuracy of VQA on VQAv2 dataset. Methods with an asterisk (*) indicate zero-shot performance, while the others show fine-tuning results. \textbf{Bottom row}, recall@10 of video retrieval on MSRVTT and YouCook2 datasets, and accuracy of action recognition on HMDB dataset. For the video retrieval tasks, asterisks denote fine-tuning performance, while non-asterisked methods denote zero-shot performance. The opposite holds true for the action recognition task. We have also included GPT-4 zero-shot performance for VQAv2 and HMDB datasets for reference.}
\label{fig:results} 
\end{figure*}

This subsection discusses popular vision-language downstream tasks, including cross-modal retrieval, visual question answering, visual reasoning, and captioning. We also present an overview of the state-of-the-art models through the lens of performance, pretrained datasets, and model parameters, which are summarized in Fig.~\ref{fig:results}.

\subsubsection{Cross-modal Retrieval}
\keypoint{Image-text Retrieval}
Image-text retrieval can take two forms: image-to-text and text-to-image retrieval which correspond to using images and text as the query modality respectively. 
The evaluation metric used to measure performance is Recall@K, with K typically set to 1, 5, or 10. Popular evaluation datasets include COCO \cite{Lin2014MicrosoftCC} and Flickr30K \cite{plummer2015flickr30k}.

\keypoint{Video-text Retrieval}
Video-text retrieval mainly focuses on text-to-video retrieval, which has two subtasks: (a) retrieving a \textit{relevant video}  based on a text query, or (b) retrieving a \textit{video segment} within a given video that matches a specific text description. In both cases, the evaluation metric is recall@K. Popular datasets for (a) include MSRVTT~\cite{Xu2016MSRVTTAL}, YouCook2 \cite{Zhou2017TowardsAL}, MSVD \cite{Chen2011CollectingHP}, etc., and for (b) include  \cite{Hendricks2017LocalizingMI}, ActivityNet Captions \cite{Krishna2017DenseCaptioningEI}, etc.

\subsubsection{Visual Question Answering and Visual Reasoning}
\keypoint{Visual Question Answering (VQA)} VQA requires a model to answer a question based on an accompanying image or video. Two common settings for VQA are: 
(a) multiple-choice where the model selects an answer from a pre-defined list, and (b) open ended where the model must generate an answer without constraints.
However, to simplify the task, many works treat VQA as a classification task, which involves selecting the most frequent answers from the training set and then building an answer candidate set. The evaluation metrics can be accuracy or VQA score \cite{Agrawal2015VQAVQ}. Popular datasets for image VQA include VQAv2 \cite{Goyal2016MakingTV}, Vizwiz VQA \cite{Gurari2018VizWizGC}, VQA-CP \cite{Agrawal2017DontJA}, etc., and for video VQA include TVQA \cite{Lei2018TVQALC}, How2QA \cite{Li2020HeroHE}, TGIF-QA \cite{Jang2017TGIFQATS}, etc.

\keypoint{Visual Reasoning} Visual reasoning tasks are designed to assess a model's high-level cognitive capabilities, including spatial reasoning, logical deduction, and common sense knowledge. %
Similar to VQA, it is evaluated by accuracy or VQA score. Popular datasets include NLVR$^2$ \cite{Suhr2017ACO}, GQA \cite{Hudson2019GQAAN}, VCR \cite{zellers2019vcr}, etc.

\subsubsection{Visual Captioning} 
Captioning is a task to generate a free-form textual caption for a given image or video. The evaluation usually follows standard text generation metrics, including BLEU, METEROR, CIDEr, etc. For image captioning, the common datasets include COCO \cite{Lin2014MicrosoftCC}, Vizwiz Caption \cite{Gurari2020CaptioningIT}, TextCaps \cite{Sidorov2020TextCapsAD}, etc. For video, datasets include MSRVTT \cite{Xu2016MSRVTTAL}, YouCook2 \cite{Zhou2017TowardsAL}, MSVD \cite{Chen2011CollectingHP}, etc.

\section{Supplementary Tables}
\begin{table*}[t]
\caption{A list of algorithms with instance discrimination objectives. In the objective (\textit{obj.}) column, C stands for contrastive and M stands for matching prediction. \textit{CG/FG Pairing} refers to the coarse-grained and fine-grained input pairing, respectively. In \textit{Modalities} column, I, V, A, L, PC, and K refer to image, video, audio, language, point cloud, and keypoint, respectively.}
\begin{center}

\begin{tabular}{llclllll}
\toprule
{\bf Method} & {\bf Obj.} & {\bf CG/FG Pairing} & {\bf FG Alignment} & {\bf Encoder/Decoder} & {\bf Loss} &  {\bf Modalities}\\
\midrule
  CMC \cite{Tian2019ContrastiveMC}  & C  & Paired/\xmark & None & Spec.(w/o fus.)/\xmark & InfoNCE & RGB, depth  \\[4pt]

  AVTS \cite{Korbar2018CooperativeLO}  & C  & Paired/\xmark & None & Spec.(w/o fus.)/\xmark  & InfoNCE & V, A  \\[4pt]
  
  AVSA \cite{morgado2020learning} & C  & Paired/\xmark & None & Spec.(w/o fus.)/\cmark & InfoNCE  & V, A  \\[4pt]

  StereoCRW \cite{chen2022sound} & C  & Paired/\xmark & None & Spec.(w/o fus.)/\xmark & InfoNCE  & V, A  \\[4pt]
  
  CLIP \cite{radford2021learning}  & C  & Paired/\xmark & None & Spec.(w/o fus.)/\xmark & InfoNCE  & I, L \\[4pt]

  MMV \cite{alayrac2020self}  &  C  & Paired/\xmark & None & Spec.(w/o fus.)/\xmark & NCE + MIL-NCE  & V, L, A \\[4pt]

  MIL-NCE \cite{miech2020end}  &  C   & Mixed/\xmark & None & Spec.(w/o fus.)/\xmark & MIL-NCE  & V, L \\[4pt]

  ALIGN \cite{jia2021scaling}  &  C   & Paired/\xmark & None & Spec.(w/o fus.)/\xmark & Normalized Softmax & I, L \\[4pt]

  VATT \cite{akbari2021vatt}  &  C  & Paired/\xmark & None & Unified/\xmark & NCE+MIL-NCE & V, L, A \\[4pt]

  SLIP \cite{mu2022slip}  &  C  & Paired/\xmark & None & Spec.(w/o fus.)/\xmark & InfoNCE+NT-Xent & I, L \\[4pt]

  COOKIE \cite{Wen2021COOKIECC}  &  C  & Paired/\xmark & Implicit & Spec.(late fus.)/\xmark & InfoNCE  & I, L \\[4pt]

  CrossCLR \cite{zolfaghari2021crossclr}  &  C  & Paired/\cmark & None &  Spec.(late fus.)/\xmark & CrossCLR loss & V, L \\[4pt]

  CrossPoint \cite{Afham2022CrossPointSC}  &  C  & Paired/\xmark & None & Spec.(w/o fus.)/\xmark & NT-Xent & I, PC \\[4pt]

  CM-CV \cite{Jing2020SelfsupervisedFL}  &  C+M  & Paired/\xmark & None & Spec.(late fus.)/\xmark & Triplet Loss+CE  & I, PC \\[4pt]

  Learnable PIN \cite{Nagrani2018LearnablePC}  &  C  & Paired/\xmark & None & Spec.(w/o fus.)/\xmark & \cite{Chopra2005LearningAS} & I, A \\[4pt]

  AVID \cite{Morgado2020AudioVisualID}  &  C  & Paired/\xmark & None & Spec.(w/o fus.)/\xmark & NCE & V, L, A \\[4pt]

  FG-MMSSL \cite{Wang2021FinegrainedMS}  &  C  & Paired/\xmark & Implicit & Spec.(late fus.)/\xmark & MIL-NCE+FG-NCE & V, L, A \\[4pt]

  L$^3$-Net \cite{Arandjelovi2017LookLA}  &  M & Paired/\xmark & None &  Spec.(late fus.)/\xmark & BCE  & V, A \\[4pt]

  AVE-Net \cite{Arandjelovi2017ObjectsTS}  &  M & Paired/\xmark & Explicit &  Spec.(w/o fus.)/\xmark & BCE  & V, A \\[4pt]

  Multisensory \cite{Owens2018AudioVisualSA}  &  M  & Paired/\xmark & Explicit & Spec.(late fus.)/\xmark & BCE  & V, A \\[4pt]
  
  LLA-CMA \cite{Cheng2020LookLA}  &  M  & Paired/\xmark & Implicit & Spec.(late fus.)/\xmark & BCE  & V, A \\[4pt]

  Sound of Pixels \cite{Zhao2018TheSO}  &  M  & Paired/\xmark & Explicit & Spec.(late fus.)/\xmark & Per-pixel BCE  & V, A \\[4pt]

  Sound of motions \cite{Zhao2019TheSO}  &  M  & Paired/\xmark & Explicit  & Spec.(late fus.)/\xmark & Per-pixel BCE & V, A \\[4pt]

  Music Gesture \cite{Gan2020MusicGF}  &  M  & Paired/\xmark & Explicit & Spec.(late fus.)/\xmark & Per-pixel BCE  & V, A, K \\

\bottomrule
\end{tabular}

\end{center}
\label{table:algorithms-contrastive}
\end{table*}

\begin{table*}[t]
\caption{A list of algorithms with clustering objectives. In the objective (\textit{obj.}) column, C stands for contrastive and M stands for matching prediction. \textit{CG/FG Pairing} refers to the coarse-grained and fine-grained input pairing, respectively. In \textit{Modalities} column, I, V, A, and L refer to image, video, audio, and language, respectively.}
\begin{center}

\begin{tabular}{llclllll}
\toprule
{\bf Method} & {\bf CG/FG Pairing} & {\bf FG Alignment} & {\bf Encoder/Decoder} & {\bf Loss} &  {\bf Modalities}\\
\midrule
    XDC \cite{Alwassel2019SelfSupervisedLB}  &  Paired/\xmark &  None & Spec.(w/o fus.)\xmark & CE & V, A \\[4pt]
  SeLaVi \cite{Asano2020LabellingUV} &   Paired/\xmark &  None & Spec.(w/o fus.)/\xmark & CE & V, A \\[4pt]
  u-HuBERT \cite{hsu2022u} &   Mixed/\xmark &  None & Spec.(late fus.)/\xmark & CE & V, A \\[4pt]
  DMC \cite{Hu2018DeepMC} &   Paired/\xmark &  Explicit  & Spec.(w/o fus.)/\xmark & Max-margin & V, A \\[4pt]
  AV-HuBERT  \cite{shi2021learning} &   Paired/\xmark &  None  & Spec.(late fus.)/\xmark & CE & I, L \\[4pt]

\bottomrule
\end{tabular}

\end{center}
\label{table:algorithms-clustering}
\end{table*}

\begin{table*}[t]
\caption{A list of algorithms with masked prediction objectives. \textit{CG/FG Pairing} refers to the coarse-grained and fine-grained input pairing, respectively. In \textit{Modalities} column, I, V, A, L, PC, and KG refer to image, video, audio, language, and knowledge graph, respectively.}
\begin{center}

\begin{tabular}{llclllll}
\toprule
{\bf Method} & {\bf AE/AR} & {\bf Inter-/intra-MP} &  {\bf CG/FG Pairing} & {\bf FG Alignment}& {\bf Encoder/Decoder} & {\bf Loss}  &  {\bf Modalities}\\
\midrule
  
  VideoBERT \cite{Sun2019VideoBERTAJ} &  AE  & \cmark/\cmark & Mixed/\xmark & None & Unified/\cmark & CE  & V, L \\[4pt]

  selfDoc \cite{Li2021SelfDocSD} &  AE  & \cmark/\cmark & Paired/\xmark & Implicit & Spec.(late fus.)/\xmark & Smooth L1  & I, L \\[4pt]

  BEiT-3 \cite{Wang2022ImageAA} &  AE  & \cmark/\cmark & Mixed/\xmark & Implicit & Unified/\xmark & CE  & I, L \\[4pt]

  Unified-IO \cite{Lu2022UnifiedIOAU} &  AE  & \cmark/\cmark & Mixed/\xmark & None & Unified/\cmark & CE  & I, L \\[4pt]

  VL-BEiT \cite{Bao2022VLBEiTGV} &  AE   & \cmark/\xmark &  Mixed/\cmark & Implicit & Unified/\xmark & CE  & I, L \\[4pt]

  VATLM \cite{Zhu2022VATLMVP} &  AE  & \cmark/\cmark & Mixed/\xmark & None & Spec.(late fus.)/\cmark & CE  & V, L, A \\[4pt]

  MAG \cite{Rahman2020IntegratingMI}  &  AE/AR  & \cmark/\xmark & Paired/\xmark & None & Spec.(late fus.)/\cmark & CE  & I, L, A \\[4pt]

  SimVLM \cite{Wang2021SimVLMSV}  &  AR  & \cmark/\xmark & Paired/\xmark & None & Unified/\cmark & PrefixLM  & I, L \\[4pt]

  Pix2struct \cite{lee2022pix2struct}  &  AE  & \cmark/\xmark & Paired/\xmark & Implicit & Unified/\cmark & CE  & I, L, HTML \\[4pt]
  
  OPT \cite{Liu2021OPTOP} &  AE+AR  & \cmark/\cmark & Paired/\cmark & Implicit & Spec.(late fus.)/\cmark  & CE  & I, L, A \\[4pt]
  
  U-VisualBERT \cite{Li2021UnsupervisedVP} &  AE  & \cmark/\cmark & Unpaired/\cmark & None & Unified/\xmark & CE  & I, L \\[4pt]
  
  VLM \cite{Xu2021VLMTV} & AE  & \cmark/\cmark & Mixed/\cmark & None & Unified/\xmark & CE  & V, L, A \\[4pt]
  
  ERNIE \cite{Zhang2019ERNIEEL} & AE+AR  & \cmark/\cmark & Paired/\cmark & None & Unified/\xmark & CE  & L, KG \\[4pt]
  
  Dragon \cite{yasunaga2022deep} & AE  & \cmark/\cmark & Paired/\cmark & None & Spec.(late fus.)/\xmark & CE+KG triplet  & L, KG \\[4pt]

\bottomrule
\end{tabular}

\end{center}
\label{table:algorithms-masked}
\end{table*}

\begin{table*}[t]
\caption{A list of algorithms with hybrid objectives. \textit{CG/FG Pairing} refers to the coarse-grained and fine-grained input pairing, respectively. In \textit{Modalities} column, I, V, A, L, and C refer to image, video, audio, language, and code, respectively.}
\begin{center}

\begin{tabular}{llllllll}
\toprule
{\bf Method} & {\bf Objective} & {\bf CG/FG Pairing} & {\bf FG Align.} & {\bf Encoder/Decoder} & {\bf Loss} &  {\bf Modalities}\\
\midrule
  MCN \cite{Chen2021MultimodalCN} & ID(C)+Cluster. & Paired/\xmark & None & Spec.(late fus.)/\xmark &   \cite{Ilharco2019LargeScaleRL}+L2 & V, L, A \\[4pt]
  
  self-detector \cite{Afouras2021SelfsupervisedOD} & ID(C)+Cluster. & Paired/\xmark & Explicit &  Spec.(w/o fus.)/\xmark & NCE+CE  & V, L \\[4pt]
  
  MDA \cite{Hong2015MultimodalDA} & Cluster.+MP(AE) & Paired/\xmark & None & Spec.(w/o fus.)/\xmark & L2  & I, Pose \\[4pt]

  UNITER \cite{Chen2019UNITERUI}  & ID(M)+MP(AE) & Paired/\cmark & Implicit & Unified/\xmark & CE+KLD  & I, L \\[4pt]
  
  ViLBERT \cite{Lu2019ViLBERTPT}  & ID(M)+MP(AE) & Paired/\cmark & Implicit & Spec.(late fus.)/\xmark & CE+KLD  & I, L \\[4pt]
  
  Oscar \cite{Li2020OscarOA}  & ID(M)+MP(AE) & Paired/\cmark & Implicit &  Unified/\xmark & CE  & I, L \\[4pt]

  UNIMO \cite{li2020unimo} & ID(C)+MP(AE+AR) &  Mixed/\cmark & None &  Unified/\cmark & CE  & I, L \\[4pt]

  VLMixer \cite{wang2022vlmixer} &  ID(C)+MP(AE)  & Unpaired/\cmark & Implicit & Unified/\xmark & InfoNCE+CE  & I, L \\[4pt]
  
  $\mu$-VLA \cite{Zhou2022UnsupervisedVP} &  ID(M)+MP(AE)  & Unpaired/\cmark & Implicit & Unified/\xmark & CE+L2  & I, L \\[4pt]
  
  ALBEF \cite{Li2021AlignBF} &  ID(C+M)+MP(AE)  & Paired/\xmark & Implicit & Spec.(late fus.)/\xmark & CE+NT-Xent  & I, L \\[4pt]
  
  FLAVA \cite{Singh2021FLAVAAF}  & ID(C+M)+MP(AE) & Mixed/\xmark & Implicit & Spec.(late fus.)/\xmark & CE+InfoNCE  & I, L \\[4pt]
  
  VLMO \cite{Wang2021VLMoUV} & ID(C+M)+MP(AE) & Mixed/\xmark & Implicit & Unified/\xmark & InfoNCE+CE  & I, L \\[4pt]
  
  BLIP \cite{Li2022BLIPBL}  & ID(C+M)+MP(AR) &  Mixed/\xmark & Implicit & Spec.(late fus.)/\cmark & CE+NT-Xent  & I, L \\[4pt]

  ActBERT \cite{Zhu2020ActBERTLG} & ID(M)+MP(AE) & Paired/\cmark & Implicit & Unified/\xmark & CE+KLD  & V, L \\[4pt]
  
  UniVL \cite{Luo2020UniViLMAU} & ID(C)+MP(AE+AR) & Paired/\xmark & Implicit & Spec.(late fus.)/\cmark & MIL-NCE+CE+NCE  & V, L \\[4pt]
  
  HERO \cite{Li2020HeroHE} & ID(M)+MP(AE) & Paired/\xmark & Implicit & Spec.(late fus.)/\xmark & CE+L2  & V, L \\[4pt]

  MERLOT Reserve \cite{Zellers2022MERLOTRN} & ID(C)+MP(AE) & Paired/\xmark & None & Unified/\xmark & CE+KLD  & V, L, A \\[4pt]

  CAV-MAE \cite{gong2023contrastive} & ID(C)+MP(AE) & Paired/\xmark & Implicit & Spec.(late fus.)/\cmark & InfoNCE+L2  & V, A \\[4pt]

  MAViL \cite{Huang2022MAViLMA} & ID(C)+MP(AE) & Paired/\xmark & Implicit & Spec.(late fus.)/\cmark & InfoNCE+L2  & V, A \\[4pt]
  
  SkillNet \cite{dai2022one}  & ID(C)+Cluster.+MP(AE) & Mixed/\xmark & Implicit  & Unified/\xmark & InfoNCE+CE  & V, I, A, L, C \\[4pt]
  
\bottomrule
\end{tabular}

\end{center}
\label{table:algorithms-hybrid}
\end{table*}

\newpage

\ifCLASSOPTIONcaptionsoff
  \newpage
\fi

\end{document}